\documentclass{article}

\usepackage{PRIMEarxiv}

\usepackage[utf8]{inputenc} 
\usepackage[T1]{fontenc}    
\usepackage{hyperref}       
\usepackage{url}            
\usepackage{booktabs}       
\usepackage{amsfonts}       
\usepackage{nicefrac}       
\usepackage{microtype}      
\usepackage{lipsum}
\usepackage{fancyhdr}       
\usepackage{graphicx}       
\graphicspath{{media/}}     
\usepackage{xcolor}
\usepackage{gensymb}
\usepackage{adjustbox}
\usepackage{epsfig}
\usepackage{caption}
\usepackage{multirow}
\usepackage{subcaption}
\usepackage{amsmath}
\usepackage{algorithm}
\usepackage{algorithmic}
\usepackage{amssymb}
\usepackage{enumitem}
\usepackage{float}
\usepackage{placeins}
\title{GCI-ViTAL: Gradual Confidence Improvement with Vision Transformers for Active Learning on Label Noise
}

\author{
  {Moseli Mots'oehli \hspace{12mm} Kyungim Baek}\\
  Department of Information and Computer Sciences\\
  University of Hawai'i at Manoa\\
  Honolulu, HI 96822 \\
  \texttt{\{moselim, kyungim\}@hawaii.edu}
}

\begin{document}
\maketitle

\begin{abstract}
Active learning aims to train accurate classifiers while minimizing labeling costs by strategically selecting informative samples for annotation. This study focuses on image classification tasks, comparing AL methods on CIFAR10, CIFAR100, Food101, and the Chest X-ray datasets under varying label noise rates. We investigate the impact of model architecture by comparing Convolutional Neural Networks (CNNs) and Vision Transformer (ViT)-based models. Additionally, we propose a novel deep active learning algorithm, GCI-ViTAL, designed to be robust to label noise. GCI-ViTAL utilizes prediction entropy and the Frobenius norm of last-layer attention vectors compared to class-centric clean set attention vectors. Our method identifies samples that are both uncertain and semantically divergent from typical images in their assigned class. This allows GCI-ViTAL to select informative data points even in the presence of label noise while flagging potentially mislabeled candidates. Label smoothing is applied to train a model that is not overly confident about potentially noisy labels. We evaluate GCI-ViTAL under varying levels of symmetric label noise and compare it to five other AL strategies. Our results demonstrate that using ViTs leads to significant performance improvements over CNNs across all AL strategies, particularly in noisy label settings. We also find that using the semantic information of images as label grounding helps in training a more robust model under label noise. Notably, we do not perform extensive hyperparameter tuning, providing an out-of-the-box comparison that addresses the common challenge practitioners face in selecting models and active learning strategies without an exhaustive literature review on training and fine-tuning vision models on real-world application data.
\end{abstract}


\section{Glossary}\label{Glossary}
Deep Learning (DL), Active Learning (AL), Deep Active Learning (DAL), Multi-Head Self-Attention (MHSA), Convolutional Neural Network (CNN), Vision Transformer (ViT), Large Language Models (LLMs), Multi-Layer Perceptron (MLP).

\section{Introduction}\label{sec:introduction}
While most works in literature often compare results to baseline active learning algorithms such as random selection or simple entropy-based selection, the extensive hyper-parameter tuning performed during training but often left out of the manuscripts leads to authors stating vastly different performances for the same CNN architecture, active learning algorithm \cite{Ren:DALSurvey20}, and label noise rate \cite{Algan:ImageNoisySurvey}. This not only raises questions about the credibility of reported state-of-the-art results but can also delay actual progress in developing active learning schemes that are robust to label noise and achieve performances comparable to models trained on clean labels. The work on non-active training of DL models in the presence of label noise, as well as the training of DL models on noise-free datasets in an active learning setting, are well addressed in the literature. However, the intersection of these niches has a long way to go \cite{Mots'oehli:DeepActiveLabelNoise23}. AL algorithms seek to train an optimal model with minimal training data that is labeled iteratively. 

\begin{figure}[!tbp]
	\begin{center}
		\includegraphics[width=1.0\columnwidth]{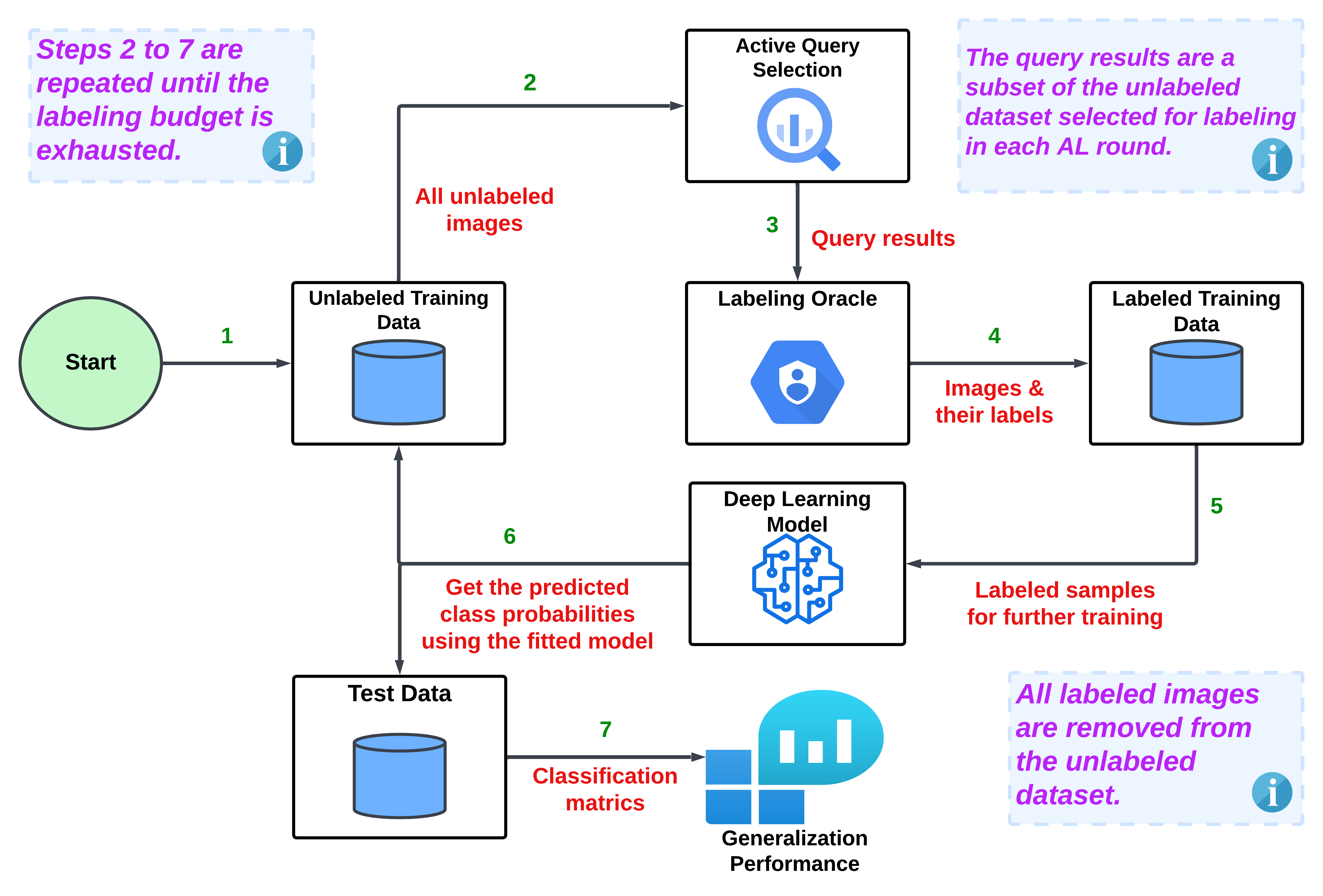}
	\end{center}
\caption{The main components in the AL framework in the presence of a noisy oracle. Each of these components may vary depending on the complexity of the data to be learned and the available resources. Most work in active learning with label noise has focused on the development of query selection algorithms that lead to highly informative and diverse data samples as well as noise-robust DL models.}
\label{fig:ALFramework}
\end{figure}

The most common AL methods seek to explore diverse training examples or focus on samples that the DL algorithm is uncertain about. AL in the presence of label noise is a particularly challenging topic since training DL models with a higher concentration of incorrect labels presents problems for the back-propagation algorithm's ability to converge as demonstrated in \cite{Algan:ImageNoisySurvey}. It has also been shown that without sufficient training data, large models can memorize the noisy labels, and fail to generalize to the test set \cite{Zhang:NoiseMemorization16, Malach:Decoupling17, Han:Coteaching18, Liu:EarlyLearningMemorization20, Liang:NoisySurvey22}. Figure \ref{fig:ALFramework} depicts the basic AL framework for training DL image classifiers in the presence of label noise.

In this work, we compile a unified view of existing works on active learning for image classification with label noise. We are particularly interested in the fine-tuning CNN and ViT models pre-trained on the imageNet-1k dataset. We explore different DL model architectures and AL strategies on different datasets while varying the label noise rates up to 60\%. We re-implement the commonly used baseline AL strategies namely: random query, maximum entropy, margin-based selection, model delta, and hybrid uncertainty sampling with diversity. 

This work seeks to address the following:

\begin{itemize}
  \item{In the realm of active learning, where the emphasis is often on query selection, and in image classification with label noise, which is typically trained without active learning, reliable benchmarks for active learning algorithms with label noise are scarce. To tackle this issue, we conduct experiments by training multiple deep-learning models for image classification using different active learning algorithms and varying levels of label noise. We present our findings by reporting test results on popular datasets, including CIFAR10, CIFAR100, Food101, and Chest X-ray images (pneumonia).}
  
  \item{Given that the ViT-based models currently outperform CNNs in image classification on CIFAR10 \cite{Kolesnikov:ViT21,Touvron:GoingDW21}, and post competitive results on CIFAR100, Food101, Chest X-ray images (pneumonia), and other classification datasets \cite{Ridnik:ImageNer21ForThemasses21,Touvron:GoingDW21,Yuan:IncorporatingCD21}, how does the ViTs compare to CNN-based models in an active learning setting with label noise (ALLN), and what can be done to improve on ViT learners in this setting?}
  
  \item{Lastly, we propose an active learning scheme customized for the properties of the transformer network to improve on active learning in the presence of label noise. We also provide new insights based on using ViTs for AL under label noise and propose avenues to advance this work.}
\end{itemize}

\section{Related Works}\label{sec:literature}
In this section, we highlight the current state of the literature on active learning, as well as active learning with label noise. We also highlight the use of the ViT for image classification and briefly discuss works that employ the ViT for active learning and label noise settings.

\subsection{Deep Active Learning with Label Noise}\label{sec:DALLN}
In most supervised machine learning use cases, there is an initial data collection and labeling cost, in both money and time. In some domains and tasks, datasets are inherently difficult to label for a variety of reasons, meaning more time is needed even by an expert human annotator to assign a label to each sample. In other cases the cost of hiring expert annotators is high, such as is the case in medical imaging \cite{Grriz:CostEffectiveA17,Konyushkova:LearningAL17}, or the cost of producing the samples is high, such as is the case in experimental physics where observations come from costly telescopes or particle accelerators with limited access \cite{Kremer:BigUbi17,Carena:NuclearInstruments14,Gutleber:HighEnergy03}. This challenges the real-world use of machine learning systems, especially as unlabeled dataset sizes increase. While much progress has been made in improving self-supervised learning (SSL) methods to leverage large unlabeled datasets for extracting quality image embeddings \cite{Zhai:S4LSS19,Chen:ContrastiveVR20,chen:SSViT21,He:MaskedAE22,assran:Ijepa23} to be used on downstream tasks with little labeled data, these methods still fall short when directly applied to datasets with noisy labels in an active learning setting \cite{Mots'oehli:DeepActiveLabelNoise23}. Active learning is a machine learning paradigm, as depicted in Figure \ref{fig:ALFramework}, that seeks to address the issues related to training ML models within a labeling budget, letting learning algorithms iteratively select a subset $L^{m}$ of size $m$, from a larger unlabelled dataset $U^n$ of size $n: m \leq n$, to be labeled by an oracle $O$ for training \cite{Lewis:PoolBased94,Cao:BALD21,Gal:BALD17,Sener:CoreSetActiveL18,HarPeled:MaxMarginCoreSet07}. However, the oracle may not always provide the correct label \cite{Younesian:ActiveLNoisyStrams20,Wei:NoisyAnnotation22,Chao:ThreeTeaching24}. The active learning mantra under label noise can be stated as follows: Train a machine learning model on a significantly smaller dataset that may contain $p\%$ label noise, with little to no drop in test performance, while staying within a pre-determined labeling budget $B$.

Most work in literature use random query, uncertainty sampling, and entropy-based sampling as baseline algorithms in comparing more complex methods for ALLN such as  \cite{Gupta:NoisyBA20, Younesian:ActiveLNoisyStrams20}, where a mixture of information gain and uncertainty is used for query selection. Other works in the literature focus on reducing the labeling budget by using a mixture of weak and strong annotators as well as annotator abstention in the case of uncertainty \cite{Younesian:ActiveLNoisyStrams20,Amin:InducedAbstention21,Younesian:QActor21}. Despite these works posting promising results in budget optimization, there is little to no improvement in terms of the robustness to label noise and improved query selection. In \cite{Huang:OracleEpiphany16}, Huang et al. show that DAL is viable with oracle epiphany, which is to say the oracle is allowed to abstain from labeling samples they are unsure about until later on in the DAL cycle once they have seen enough examples to provide a more confident label. While their method is more realistic and leads to better performance, abstention may not always be possible in a fast-paced sector where lots of data is generated on a daily basis and requires urgent labeling. In \cite{Yan:ImperfectLabelers16}, Yan et al. utilize abstention in DAL under label noise. A key difference in their work is that the algorithm need not be aware of either the abstention or noise rate. While there are obvious merits to the methods above, the solutions rely too heavily on the specific setup of the AL cycle and the human annotators to be trusted in the general setting. For these reasons and more, our work focuses on a more algorithmic approach to robustness and query selection. 

Recent work \cite{Roschewitz:ALDatasetQuality22} deploys a more data-centric approach to label noise for active label cleaning by ranking samples for label correctness and labeling difficulty. In \cite{sheng2024foster}, the authors propose a data-driven self-adapting DAL strategy that selects potentially noisy labels for correction in a manner that automatically avoids class imbalance in the labeled dataset with no prior knowledge of the class distributions. A thorough study of the literature \cite{Ren:DALSurvey20,Mots'oehli:DeepActiveLabelNoise23,li:empiricalEfficacy22} raises questions on the superiority and robustness of more complex DAL methods. The authors state that due to a lack of standardized benchmark settings in overlapping niches, performances by baseline models and AL queries in noisy labels tend to be understated. The use of grid search and related methods for hyper-parameter tuning can also aid in finding the ideal model parameters that show the superiority of a proposed AL model over the baseline methods. It is for this reason we opt to avoid any extensive hyper-parameter optimization that could skew the results of this study, and thus we report a realistic out-of-the-box benchmark in AL under label noise.

\subsection{Vision Transformer for Image Classification}\label{sec:ViTClassifcation}

\begin{figure}[!bp]
	\begin{center}
		\includegraphics[width=1.0\columnwidth]{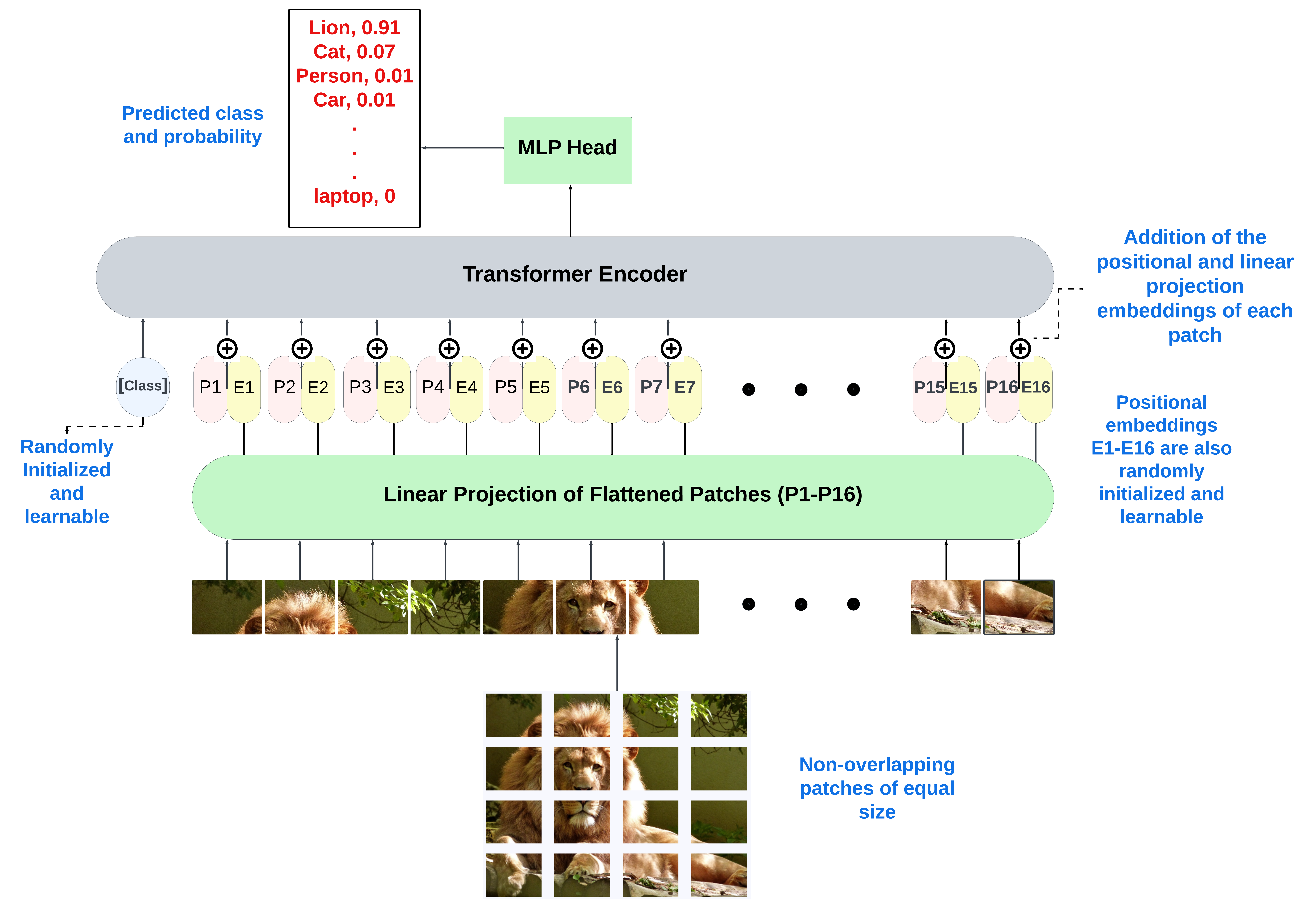}
	\end{center}
\caption{ViT architecture showing how image patches are extracted as well as their positional embedding. The transformer encoder can contain multiple attention and normalization layers. Finally, fully connected layers are added, with a softmax operation for image classification. (Adopted from \cite{Kolesnikov:ViT21})}
\label{fig:ViT}
\end{figure}

In \cite{Kolesnikov:ViT21}, Kolesnikov et al. present the Vision Transformer (ViT) as a CNN replacement for image classification tasks. They show that, in the extensive dataset regime, ViTs achieve higher classification accuracy, are more computationally efficient, and show no signs of saturation compared to CNNs such as ResNet and EfficientNet on increasingly larger datasets. The standard ViT architecture takes 16 by 16 patches from an image, flattens them, and applies a linear projection onto a higher dimensional space equal to that of the original text-based transformer. The patches are then marked for where in the image they were extracted, and so a second input to the transformer is the 1D positional embedding of each patch as shown in Figure \ref{fig:ViT}. The spatial correlations of the patches are learned implicitly through their positional embeddings and self-attention vectors. The positional embeddings ensure the model learns both the relationships between pixels in the patches as well as the local and global 1D proximity representations of the tokens (image patches). Below we give an overview of the main transformer encoder and the attention mechanism within the ViT architecture.

\begin{figure}[!bp]
	\begin{center}
		\includegraphics[width=1.0\columnwidth]{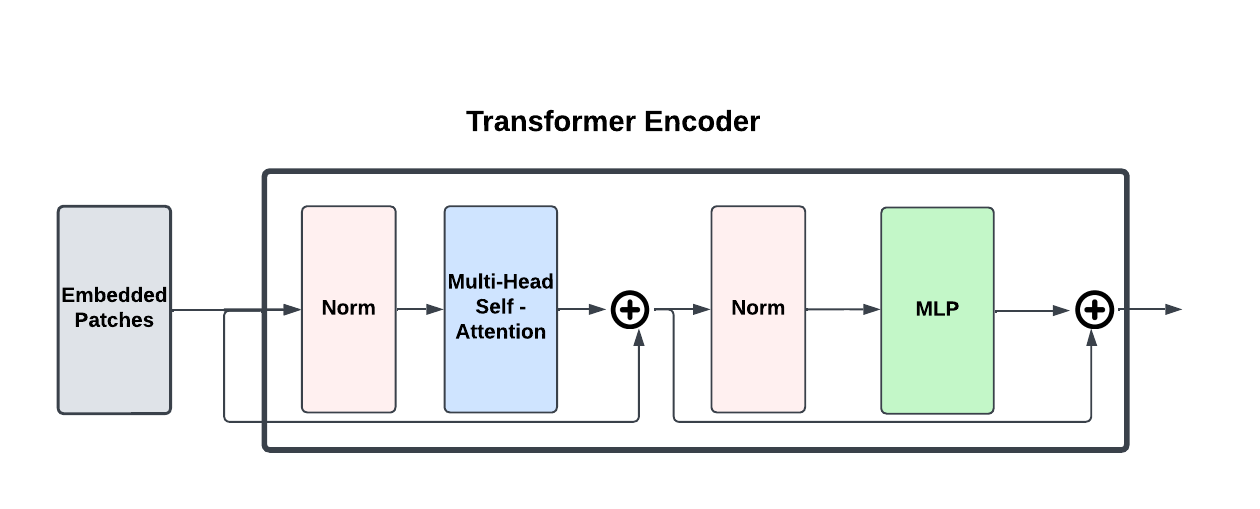}
	\end{center}
\caption{The transformer encoder block that constitutes the main component of representation learning in both large language models (LLMs) and ViTs. The main transformer components are the Multi-Head Self-Attention encoder blocks and the Multi-Layer Perceptron (MLP) layers. The normalization layer helps ensure the model is robust to covariate shifts in the features within a batch. (Adopted from \cite{Kolesnikov:ViT21})}
\label{fig:transformer_block}
\end{figure}

\textbf{Tansformer Encoder: } The transformer encoder block in a ViT consists of multi-head self-attention (MHSA) and multi-layer perceptron (MLP) layer with normalization layers before and after MHSA each performing specific operations on the input tokens or subsequent later outputs. MHSA attempts to learn diverse features to capture semantical and contextual complexity in image patches while learning all these features in parallel for a given input. The parameterized MLP pools and aggregates the learned high-dimensional features from all the last layer attention heads, and compresses them into a lower-dimensional representation for a downstream task such as image classification, regression, and more. Figure \ref{fig:transformer_block} illustrates the transformer encoder block architecture.

\textbf{Self-Attention: } Given a sequence of input vectors, the self-attention mechanism calculates the similarity between the vectors. In the case of ViTs, starting with a sequence of $m$ patch embeddings of size $e$,  $\mathbf{X}= [x_{1}, x_{2}, ...,x_{m}]$, where $x_{i} \in \mathbb{R}^{e}$ and so $\mathbf{X} \in \mathbb{R}^{m \times e}$, the goal of the self-attention mechanism is to learn a representation of each input token, expressed as a vector that is a weighted sum of the other token representations, where the weighting is based on how similar the tokens are to each other contextually. This similarity is measured by taking the dot product of any two embedding vectors, representing the cosine similarity between two vectors. To achieve this, three parameterized matrices are used: the Query matrix $\mathbf{W}^{Q} \in \mathbb{R}^{e \times e_{q}}$, the key matrix $\mathbf{W}^{K} \in \mathbb{R}^{e \times e_{k}}$, and the value matrix $\mathbf{W}^{V} \in \mathbb{R}^{e \times e_{v}}$, such that $e_{q} = e_{k}$. To calculate the attention weights, the input sequence $\mathbf{X}$ is first projected into the query, key, and value representations through matrix multiplication to get $\mathbf{Q} = \mathbf{XW}^{Q}$, $\mathbf{K} = \mathbf{XW}^{K}$, and $\mathbf{V} = \mathbf{XW}^{V}$. Applying a softmax operation to the normalized dot product of the query and key projections of the input produces the attention weights. The final output of the attention layer is a weighted sum of the value representations, where the weights are the attention weights:

\begin{equation}\label{eq:attention_vectors}
    \mathbf{Z} = softmax \Bigg(\frac{\mathbf{QK}^{T}}{\sqrt{e_{q}}}\Bigg)\mathbf{V}
\end{equation}

\textbf{Multi-Head Self-Attention (MHSA):} Extends the capabilities of self-attention by allowing the model to use multiple self-attention heads with different learned parameters in parallel over the same input. Learning using one attention would force the model to compress all learned information about an input image into one fixed-length vector. Using multiple attention heads allows the model to learn and express semantic and contextual complexity over several specializing attention heads simultaneously. A ViT can have one or more MHSA encoders with residual connections to maintain lower-level feature information throughout the network for context in higher-level attention blocks. In this work, we focus on the last layer of MHSA heads as well as the predicted class probabilities after the MLP is trained for classification.

Figure \ref{fig:ViT} is an illustration of the ViT architecture, showing patch and positional embeddings, as well as the transformer encoder adapted from \cite{Kolesnikov:ViT21}. Other ViT variations that achieve performances comparable to and in some cases better than CNNs on image classification tasks include Swin Transformer \cite{liu:swin21}, Transformer in Transformer (TNT) \cite{Han2021TransformerIT}, DaViT-G \cite{Ding:DaViT22}, and Multiscale Vision Transformers (MViT) \cite{fan:multiscale21}. The Swin transformer adapts a similar overall structure as the ViT but uses shifted overlapping patches instead of rigid non-overlapping patches. The authors of \cite{liu:swin21} argue and demonstrate that using shifted window patches instead of non-overlapping patches limits the self-attention operations to locally related patches of an image. This reduces the complexity of the self-attention operations from quadratic to linear on the input. This architectural choice makes the Swin transformer more efficient than the original transformer while performing comparably in terms of top-1 classification accuracy. That said, the performance of ViTs and their variants such as the Swin transformer compared to CNNs has not received much attention in the noisy label domain or AL. For these reasons, we included both the base ViT and the Swin transformer in our experiments. 

Previous works that adopt the ViTs for image classification in the AL domain include \cite{Caramalau:VisualTF2021,HE:ViTMedical21}.  Caramalau et al. \cite{Caramalau:VisualTF2021} introduce a novel AL query strategy that combines CNN layers for local dependencies and ViTs to capture non-local dependencies while jointly minimizing a task-aware objective. They achieve state-of-the-art performance on most AL-based benchmarks. Their method however suffers from scalability limitations due to ViT's large parameter space and potential batch size restrictions in training. A similar conclusion is reached by He et al. \cite{HE:ViTMedical21}. Their work demonstrates that, while ViTs produce informative and task-aware AL queries on CIFAR10 and 100, they are considerably larger than CNNs in terms of model parameters for them to be a viable replacement in DAL with the existing hardware in terms of training time. The work of Rotman and Reichart \cite{Rotman:MultiTaskALTransformerBased22} compares different DAL methods on different text classification datasets using transformer-based models. While their work is not focused on image classification or the vision transformer, they demonstrate that transformer-based models tend to lead to inconsistent and poor results in the AL setting when using basic AL strategies. They show that query selection based on a transformer learner sometimes leads to the selection of clusters of neighboring outliers that destabilize training. 

In this work, we introduce a novel DAL algorithm, GCI-ViTAL, that takes advantage of the ViT's ability to capture complex local and global dependencies like \cite{Caramalau:VisualTF2021}. We define the C-Core attention vectors to help reduce the computational complexity for comparisons between labeled and unlabeled samples per AL circle. Like \cite{HE:ViTMedical21}, we make use of the C-Core attention vectors as an informative and noise-aware approach to sample selection.

\section{Query Based on ViT Patch Similarity}\label{Chapter_4}
Most AL approaches rely solely on the predicted probabilities from the trained model to form their query strategy. Label noise leads to a confused learner that outputs uncalibrated probability estimates of the samples, so selecting samples based on the predicted probabilities alone leads to the selection of non-optimal samples for each iteration and further corruption of the model through unstable gradients from incorrectly labeled samples. To address this issue, our proposed solution involves using the attention vectors from the last layer of a ViT model in query selection. This approach is based on finding specific features of a ViT that can help identify potentially mislabeled samples by comparing the labels of image samples that are close to each other in the last layer of attention representation maps. We start with a pre-trained ViT and fine-tune it using a small set of images with accurate labels, a common practice in many AL algorithms. Subsequently, we extract attention vectors for all the images in the initial set with accurate labels, and we use these to create core attention representations (centroids) for each class by aggregating the attention vectors of images belonging to the same class. This results in C-Core attention representation vectors that have been trained exclusively using accurate data for a C-class classification problem. The purpose of these core attention vectors is to help identify potentially incorrect labels during the AL process and reduce their impact on the model's training. In each AL cycle, unlabeled images go through the fine-tuned ViT model, and their last layer's attention vectors are collected. Our AL strategy combines the consideration of prediction uncertainty and attention-based diversity, which means we select samples that the current model is uncertain about while also striving to maintain a balanced representation of each class in the training dataset using the C-Core attention vectors.

\subsection{Handling Label Noise through Gradual Class-Centric Confidence Improvement}
During training, an oracle receives a batch of $K$ samples to label, and before retraining the model with these labeled samples, a portion of them that deviate too much from the C-Core attention vector of the assigned class have their class assignment probabilities changed. We explore handling these examples in two ways, first by assigning the samples to the class dictated by the C-Core attention vectors, or, secondly by label smoothing. With label smoothing, we change the class probabilities assigned by the oracle for a sample so that it is not one-hot-encoded, but rather we introduce a positive probability for another class. Since we have the C-Core class centroids, we smooth the label by assigning a positive probability to the closest class centroid from the current sample in the attention vector space. For example, say sample $x_{j}$ was selected for labeling, and the noisy oracle assigns it to class $C_{4}$ out of 10 classes. This means the probability distribution is given by $[0,0,0,1,0,0,0,0,0,0]$. However, if we suspect this labeling to be incorrect based on the sample $x_{j}$ being far from the core attention centroid of class $C_{4}$, we alter the probability distribution in such a way that we express less confidence in class $C_{4}$ being the correct label. If the closest centroid in attention vector space is that of class $C_{9}$, we change the probability distribution $p(y|x_{j})$ to be $[0,0,0,(1-\epsilon),0,0,0,0,\epsilon,0]$, where $\epsilon \in [0,1]$ controls how much to trust the oracle as opposed to the C-Core attention vectors that compare images semantically. 

Once the initial model is trained on the clean set, future batches for labeling are selected in a way that promotes class-centric confidence. We first calculate prediction uncertainty for each class based on the model's confidence in its predictions. The AL strategy selects samples with the highest distance to their core attention class centroid, measured by the Frobenious norm. We then rank these from highest to lowest based on the Frobenious distance and take the top-$K$ for labeling. As the model improves its performance on the validation set, we shift towards selecting samples with high uncertainty and gradually decreasing reliance on the class centroid distance to the image representation, ultimately focusing on refining the model's understanding of each class.

\begin{figure}[!tbp]
	\begin{center}
		\includegraphics[width=1.0\columnwidth]{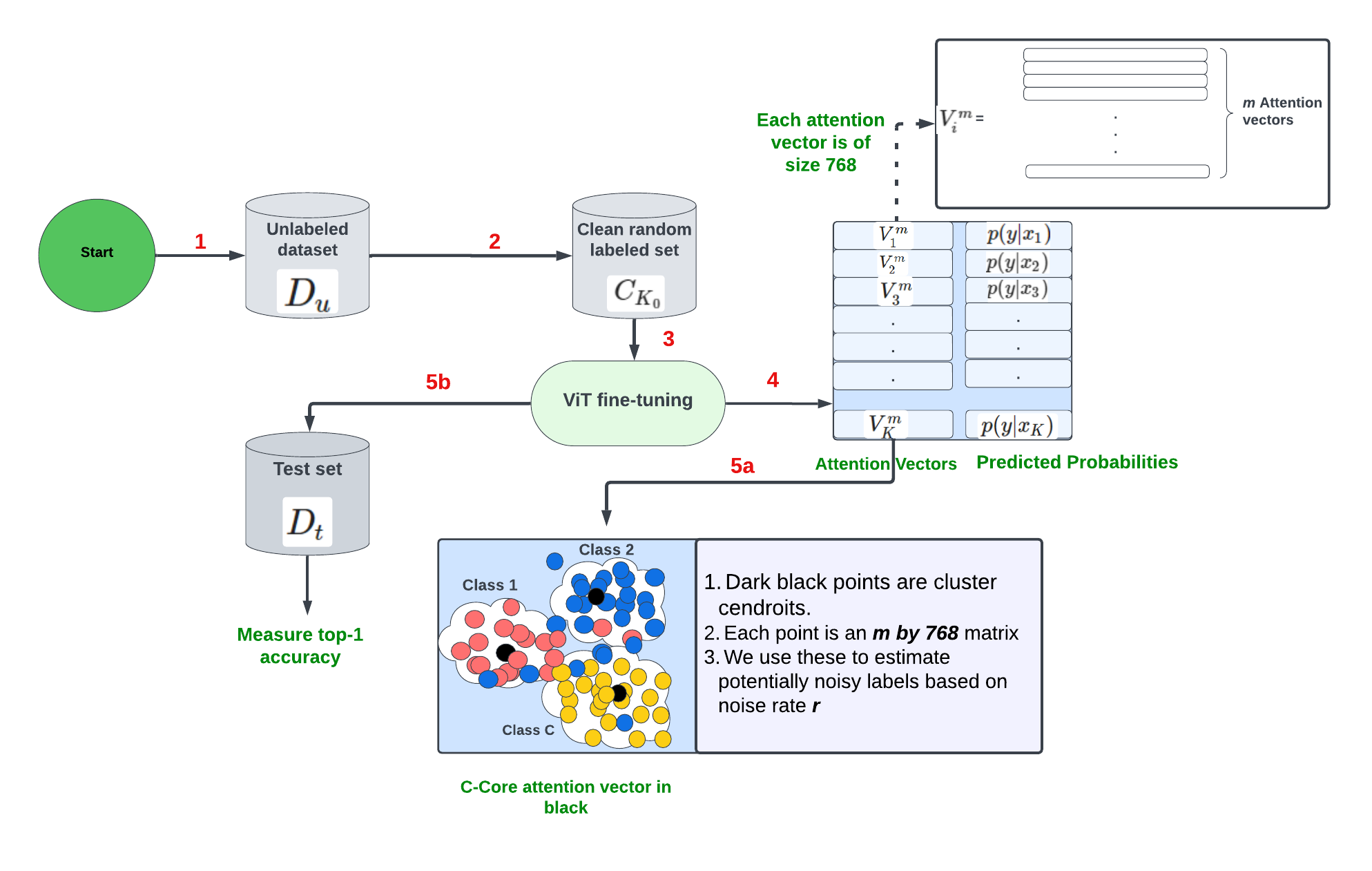}
	\end{center}
\caption{The first stage in the AL cycle where the ViT model is fine-tuned on a clean random set and the C-Core attention vectors are computed for each cluster in the clean set. Once the initial training is done, the model and C-Core vectors are iteratively used in selecting samples for labeling.}
\label{fig:ALFrameworkNewFirstStage}
\end{figure}

\begin{figure}[!tbp]
	\begin{center}
		\includegraphics[width=1.0\columnwidth]{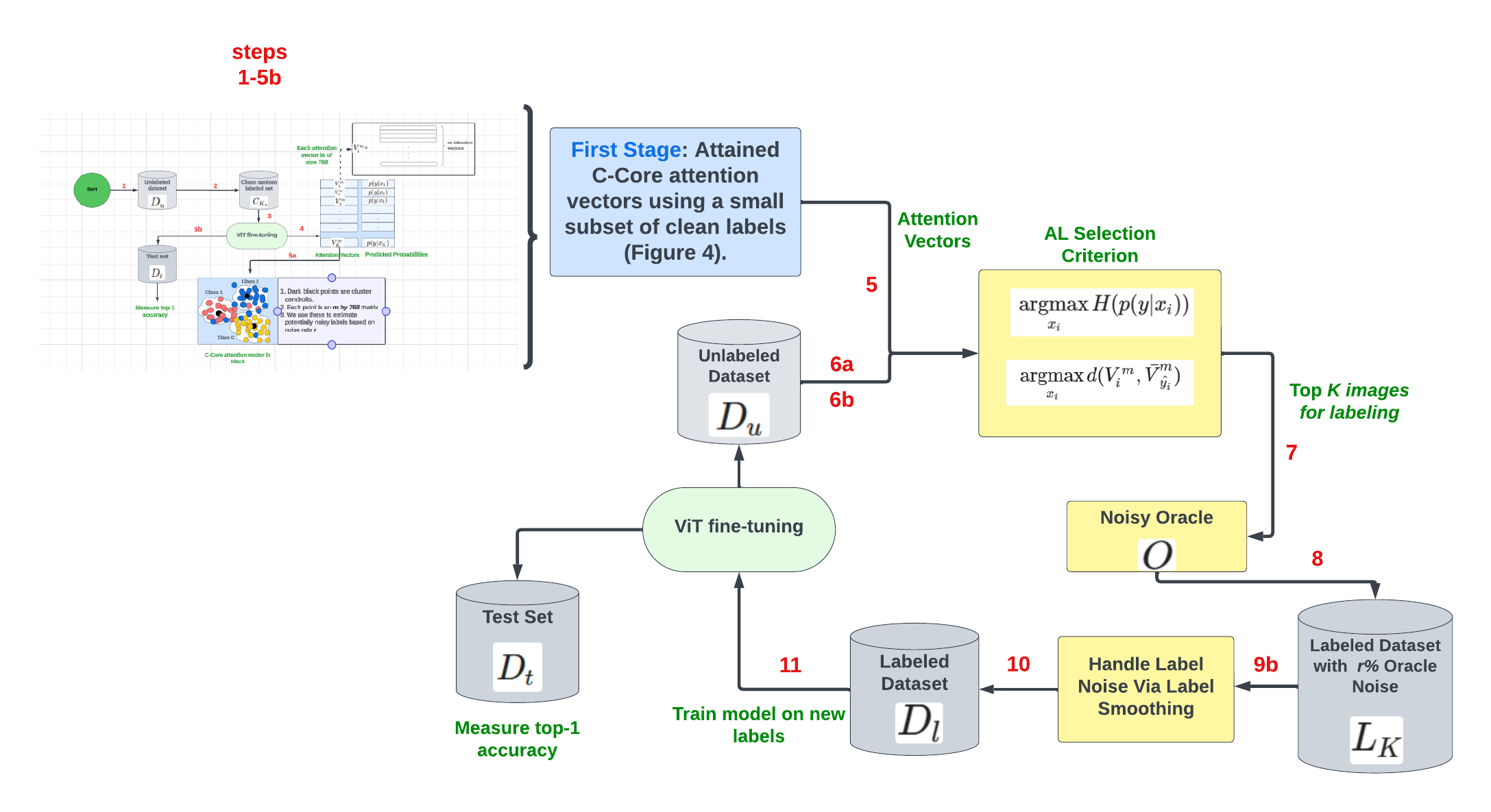}
	\end{center}
\caption{The GCI-VITAL DAL framework. This diagram shows the iterative active learning cycle, where C-Core attention vectors from the ViT model guide the selection of semantically challenging samples based on their distance from class centroids. Label smoothing mitigates noise, enhancing model robustness. Steps 6a to 11 continue until the labeling budget is exhausted.}
\label{fig:ALFrameworkNew}
\end{figure}

The selection based on the distance from the class centroid ensures we select samples that have underlying attention maps that are not similar to most seen in the clean training set for that class, which are effectively semantically hard examples. We want these to be sent to the oracle for labeling. Our proposed AL strategy for image classification in the presence of label noise by leveraging attention vectors is summarized in Figures \ref{fig:ALFrameworkNewFirstStage} and \ref{fig:ALFrameworkNew}. The detailed description of the steps in Figure \ref{fig:ALFrameworkNew} is provided in Section \ref{subsec:formulation}.

In summary:

\begin{enumerate}[label=\arabic*.]
\item Most AL strategies rely solely on model-predicted probabilities and the oracle's labels, which may lead to suboptimal sample selection and unstable gradients due to label noise.

\item Our approach uses the last-layer attention vectors (C-Core attention vectors) of the ViT, to identify potential images with semantic similarity that have opposing oracle labels. These are potentially incorrectly labeled. We reduce the effects of label noise through label smoothing by combining prediction uncertainty and the distance to the most likely C-Core attention vector.

\item We employ a Gradual Class-Centric Confidence Improvement strategy, initially selecting samples with a high Frobenius norm concerning their supposed class centroid attention maps for low-confidence classes. Gradually, we shift our strategy towards samples with a low Frobenius norm to better understand the classes. This approach helps in the selection of semantically challenging examples and reduces the impact of label noise.

\end{enumerate}

In the rest of this section, we provide the theoretical formulation of the problem, the algorithm, and the theoretical justification for our approach.

\subsection{Joint Entropy-Attention Active Learning}\label{subsec:formulation}
Starting with: (1) An initial set of $K_{0}$ labeled images, referred to as the ``clean labeled set'', consisting of randomly selected images. This set is denoted as $C_{K_{0}}$ and should be large enough to ensure the representation of at least a few examples from each of the $C$ classes. (2) A base Vision Transformer (ViT)  denoted by $ViT_{b}$. The ViT has been pre-trained on ImageNet-1k \cite{Russakovsky:ILSVRC15}, and has the fully connected layers changed to suit the number of classes $C$. (3) Iteratively fine-tune the ViT on a collection of $K$ images with the highest prediction entropy $H(p(\hat{y}|x_i))$ and ViT hidden layer attention heads distance to their supposed centroid attention heads, i.e., the images which jointly maximize Equations \ref{eq:init_entropy} and \ref{eq:init_frobenius}:

\begin{equation}\label{eq:init_entropy}
    \underset{x_i}{\text{argmax}}\, H(p(\hat{y}|x_i)) = \underset{x_i}{\text{argmax}}\big[ -\sum_{y=1}^C p(y|x_i) \log p(\hat{y}|x_i) \big]   
\end{equation}

\begin{equation}\label{eq:init_frobenius}
\underset{x_i}{\text{argmax}}\, d(V^{m}_{i}, \overline{V}^{m}_{\hat{y_i}})
\end{equation}

where $d$ is a suitable distance measure, $C$ is the number of classes, and $\hat{y_i}$ is the predicted class of $x_{i}$.  We also have: 

the $m$ attention vectors $V^{m}_{i}$ produced by running image $x_{i}$ through the fine-tuned model and

\begin{equation}\label{eq:centroid}
    \overline{V}^{m}_{\hat{y_i}} = \frac{\sum_{x_j \in C_{K_{0}}}{z_{j}^{\hat{y}_{i}}}V_{j}^{m}}{\sum_{x_j \in C_{K_{0}}}{z_{j}^{\hat{y}_{i}}}}
\end{equation}

where

\begin{equation}
    z_{j}^{\hat{y}_{i}} = \begin{cases}
1 & \text{if } x_j \in \text{class } \hat{y}_{i} \\
0 & \text{otherwise}
\end{cases}
\end{equation}

is an entry in an indicator vector, where each entry is 1 if the image $x_{j}$ belongs to class $\hat{y}_{i}$. Consequently, the denominator in Equation \ref{eq:centroid} represents the number of images in the initial clean labeled set that belong to class $\hat{y}_{i}$.


Equation \ref{eq:centroid} represents each of the C-Core attention maps or centroids for all images in the clean labeled random set belonging to the same class. Note that these class centroids are produced only after fine-tuning on this labeled set with no label noise. These represent our best estimate of the truth at this point before oracle noise influences our AL strategy. $V^{m}_{i}$ and $\overline{V}^{m}_{\hat{y_i}}$ are each a collection of $m=12$ vectors in the case of $ViT_{b}$, for each of the 12 attention heads, of embedding size of $768$ 

We will use the notation $V^{m}_{i}$ and leave out the 768 dimension going forward, however, each $V^{m}_{i} \in \mathbb{R}^{m \times 768}$ will be treated as a matrix containing attention vectors.
Calculating the distance between matrices is typically done by using the Frobenius norm $(F_{n})$ \cite{Frobenius:UeberlineareSubstitutionen1877}, the spectral distance $(S_{d})$ \cite{Kazakos:SpectralDM80}, or the Kullback-Leibler divergence $(D_{KL})$ \cite{Kullback:KLdivergence51}. $F_{n}$ is calculated by summing the squares of the differences between all the elements of the two matrices and then taking the square root of the sum, similar to the $L_{2}$ distance between vectors. $S_{d}$ distance quantifies how different two matrices $A$ and $B$ are based on the largest eigenvalue of the difference matrix between them $||A-B||_{2}$. $D_{KL}$ treats matrices as probability distributions and measures the amount of work to be done in transforming one distribution into another. Similar to the works in \cite{Guo:2019AFN, Yuan:MaximumBatchFrobeniusNorm22, Qiu:ProbabilisticNormClustering18, spatialFrobeniousJaya:Ramnarayan19} we argue the Frobenius norm between the latent representations of input images or text can improve discriminability. Unlike these works, we however do not use the Frobenious norm as a parameterized regularizer in training the network, but rather as part of our AL query strategy. Since we are comparing $m \times 768$ centroid matrices to $m \times 768$ image attention maps, the Frobenius norm also becomes a natural choice in calculating the distance in the semantically aware representations learned by self-attention. The norm is given by:

\begin{equation}\label{eq:real_frobenius}
    d\big(V^{m}_{i}, \overline{V}^{m}_{\hat{y}_i}\big)_{F_{n}} = \sqrt{\sum_{j=1}^m \sum_{k=1}^{768} (V^{jk}_{i} - \overline{V}^{jk}_{\hat{y}_{i}})^2}    
\end{equation}

To have a combined objective for optimization, we scale the distance to be in the same range as the entropy. For a two-class image classification problem, the entropy is between 0 and 1, but for a multi\-class problem, the entropy is between 0 and $log(C)$, where $C$ is the number of classes. An easy way to achieve this for a loss with range $[0,\infty]$ is to first rescale values
to the range $[0,1]$ and then scale all outputs to $\log(C)$. This would then yield a final range: $[0, log(C)]$. For image $x_{i}$, we calculate $F_{n}$, the distance to each clean set cluster centroid to get vector distances to all the $C$ classes. We then apply a softmax function that converts this into a probability distribution over the classes. This is to say, based on features alone, which class is most semantically similar to the image in question. We then multiply this probability vector by $log(C)$ to standardize it to the magnitude of the entropy selection criterion. We introduce a weight $\lambda \in[0,1]$,  that balances and controls the influence of entropy versus class-based feature similarity. The final objective is given by:

\begin{equation}\label{eq:all_the_equation_components}
    \underset{x_i}{\text{argmax}} \big[\lambda H(p(\hat{y}|x_i)) + (1-\lambda)\text{softmax}(d(V^{m}_{i}, \overline{V}^{m}_{\hat{y}_i})_{F_{n}}) \log(C)\big].
\end{equation}
 
Combining equations \ref{eq:init_entropy}, through \ref{eq:all_the_equation_components} we get the following:

\begin{equation}\label{final_loss_frobenious_C_Core_attention}
    \underset{x_i}{\text{argmax}} \big[-\lambda\sum_{y=1}^C p(y|x_i) \log p(\hat{y}|x_i) + (1-\lambda)\frac{e^{d(V^{m}_{i}, \overline{V}^{m}_{\hat{y}_i})_{F_{n}}}}{\sum_{y=1}^{C}e^{d(V^{m}_{i}, \overline{V}^{m}_{y})_{F_{n}}}} \log(C)\big]
\end{equation}

Algorithm \ref{alg:al_entropy_attention} below shows the pseudo-code for GCI-ViTAL, the AL query strategy we propose in this work in the presence of label noise.

\begin{algorithm}[H]
\caption{GCI-ViTAL: An AL strategy for handling label noise using class prediction entropy and the ViT final layer attention head vectors to identify candidate images of which the most recently trained model is uncertain, and the images seem to be semantically different from typical images in their supposed class label based on the extracted ViT attention head vectors.}
\label{alg:al_entropy_attention}
\begin{algorithmic}[1]
\REQUIRE:
  \begin{enumerate}
    \item A pre-trained ViT model $ViT_{b}$
    \item A small initial random set $C_{K_{0}}$ of $K_0$ images with accurate labels from all classes
    \item A set of unlabeled images $D_{u}$
    \item A labelling oracle $O_{r}$ with know n-symmetric label noise rate $r$
    \item A number of samples to label, $K$, per labeling cycle
    \item A weight $\lambda \in[0,1]$ to balance the influence of entropy versus class-based feature similarity
  \end{enumerate}    
  
\ENSURE
  Current labelling budget $B_{t} \ge 0$

\STATE Randomly initialize $D_{l} = C_{K_{0}}$
\STATE Fine-tune $ViT_{b}$ using labeled set $D_{l}$
\STATE Extract attention vectors for all images in $D_{l}$
\STATE Create the C-Core attention representation vectors (centroids) for each class in $D_{l}$

\STATE \textbf{while} $D_{u} \neq \emptyset$ \textbf{do}
    \STATE \hspace{1cm} Select $K$ samples from $D_{u}$ using the following strategy:
    \STATE \hspace{1.5cm} Calculate prediction entropy for each class based on the model's confidence in its predictions
    \STATE \hspace{1.5cm} Calculate the distance to each C-Core attention vector to get a distance vector to all the $C$ classes
    \STATE \hspace{1.5cm} Apply a softmax function to convert the distances to a probability distribution over the classes
    \STATE \hspace{1.5cm} Calculate the final objective for each sample:
      \begin{equation}
        \underset{x_i}{\text{argmax}} \big[\lambda H(p(\hat{y}|x_i)) + (1-\lambda)\text{softmax}(d(V^{m}_{i}, \overline{V}^{m}_{\hat{y}_i})_{F_{n}}) \log(C)\big]
      \end{equation}
    \STATE \hspace{1.5cm} Select the top-$K$ samples with the highest final objective
  \STATE \hspace{1cm} Send the $K$ selected samples to the oracle for labeling
  \STATE \hspace{1cm} Calculate disagreement between the oracle's labels and the C-Core attention-based class assignment for each of the $K$ samples
  \STATE \hspace{1cm} Apply label smoothing on all potentially noisy labels based on the C-Core attention vectors
  \STATE \hspace{1cm} Update $D_{l}$ with the labeled samples
  \STATE \hspace{1cm} Fine-tune $ViT_{b}$ using $D_{l}$
  \STATE \hspace{1cm} Measure top-1 accuracy on the test set
  \STATE \hspace{1cm} Update $D_u$ by removing the $K$ selected samples
\STATE \textbf{end while}
\STATE \textbf{return}
\end{algorithmic}
\end{algorithm}

\subsection{Theoretical Analysis}\label{subsec:theoretical_analysis}
In this section, we investigate the theoretical relationship between the predicted class probability distribution entropy, the Frobenius norm of the final layer ViT attention heads of the potential AL query candidates, and the C-Core attention head vectors. We then analyze the relationship between our strategy and increased label noise.

\subsubsection{Attention, Entropy and the Frobenius Norm}\label{subsub:AttEntFro}
Starting with the self-attention head outputs of samples in the clean initial random sample: 

\begin{equation}\label{eq:attention_equation}
    \mathbf{Z} = softmax \Bigg(\frac{\mathbf{QK}^{T}}{\sqrt{e_{q}}}\Bigg)\mathbf{V}
\end{equation}

for a $C$ class image classification problem, we first calculate the C-Core attention vectors as the mean attention vector representation of all images in each class. These are synonymous with cluster centroids. The Frobenius norm component in the query strategy in Equation \ref{final_loss_frobenious_C_Core_attention} measures the disparity between the attention heads of a given image and the C-Core attention vector of the predicted class. By doing this, part of the query selection strategy prioritizes data points where the attention heads in the final layer of the model capture information that is distinct from the C-Core attention vectors of the predicted class. Such images represent novel and unseen variations within a class based only on the attention mechanism. The entropy component of the query strategy on the other hand only uses the predicted class probabilities in ranking samples. Entropy quantifies the uncertainty in a model's predictions for a given image. Images with high prediction entropy are often not well represented in the dataset and thus selecting these samples provides more information. The combination of entropy and the Frobenius norm thus strikes a good exploration and exploitation balance of the image space. This leads to queries that gradually build a model trained on diverse data samples based on the semantical representation of the images through the use of the C-Core attention vector representation of samples, while gradually building up the model's confidence around the decision boundaries between classes based on both semantics as well as filtered label information from the oracle.

\subsubsection{Label Smoothing Using the C-Core Attention-Vectors to Reduce the Effects of Label Noise}\label{subsub:GCI-VITAL}
In this section we show how label smoothing potentially incorrect oracle labels helps reduce the adverse effects of incorrect labels during training. Mathematically, we can express label smoothing as a form of regularization or penalty-based learning. In the case of multi-class image classification using the cross-entropy loss, consider the following scenario: For a single training example $x_{i}$ with predicted probabilities $p(\hat{y}|x_{i})$, the cross-entropy loss without label smoothing is given by:

\begin{equation}\label{eq:cross_entropy}
    L = -\lambda\sum_{y=1}^C p(y|x_{i}) \log p(\hat{y}|x_{i})    
\end{equation}

where $p(y|x_{i})$ is the one-hot encoded label from the oracle, and $p(\hat{y}|x_{i})$ is the predicted probability distribution by the model. With label smoothing the cross-entropy loss is given by:

\begin{equation}\label{eq:label_smoothed_cross_entropy}
    L = -\lambda\Bigg[\sum_{y=1}^C (1-\epsilon) p(y|x_{i}) \log p(\hat{y}|x_{i}) + \frac{\epsilon}{C}\sum_{y=1}^C \log p(\hat{y}|x_{i})\Bigg]
\end{equation}

where $\epsilon$ is the smoothing parameter.

The term $-\frac{\epsilon}{C}\sum_{y=1}^C \log p(\hat{y}|x_{i})$ in the loss is a regularization term that penalizes the model for being too confident in noisy label settings, thus forcing it to learn more robust decision boundaries between classes. However, in this case, $\epsilon$ is shared between the other $C-1$ classes to distribute uncertainty amongst them equally. In our case, since we have the C-Core attention vector per class, we only label smooth by assigning the C-Core selected class a non-zero probability. This means that each sample that the oracle assigns to a class other than that we would assign using the Frobenious distance between the image's attention vectors and the C-Core vectors is label smoothed to reflect lower than $100\%$ confidence in the oracle's label, and exactly $\epsilon \%$ confidence in the class based on the distance to the C-Core attention vectors. Mathematically, we add an indicator variable to Equation \ref{eq:label_smoothed_cross_entropy} so that all other classes that are not the C-Core prediction do not get their zero probability adjusted. The cross-entropy loss based on C-Core Frobenious label smoothing is given by: 

\begin{equation}\label{eq:label_smoothed_cross_entropy_C_Core}
    L = -\lambda\Bigg[\sum_{y=1}^C (1-\epsilon) p(y|x_{i}) \log p(\hat{y}|x_{i}) + I(\hat{y}=\hat{y}_{cc})\epsilon\sum_{y=1}^C \log p(\hat{y}|x_{i})\Bigg]
\end{equation}

where

\begin{equation}
   I(\hat{y}=\hat{y}_{cc}) = \begin{cases}
1 & \text{if } \hat{y} = \hat{y}_{cc} \\
0 & \text{otherwise}
\end{cases}
\end{equation}
 
and $\hat{y}_{cc}$ represents the class that would be assigned based purely on the Frobenious norm using the C-Core attention vectors. Focusing on the terms of Equation \ref{eq:cross_entropy}, we see the loss is large when the disparity between $p(\hat{y}|x_{i})$ and $p(y|x_{i})$ is large. Let us consider two cases, a perfect model (one that can fully predict the underlying ground truth) and a random model (one that assigns random labels for any input). All other scenarios lead to a model contained in the search space of models we seek to optimize. Assuming the perfect model produces $p(\hat{y}|x_{i})$ approximately equal to the ground truth label distribution, then the loss changes based on the noise rate of the oracle. At noise rate $r=0$, the loss $L \approx 0$, and as $r \to 1 : L \to \infty$ .

Comparatively, the smoothed loss in Equation \ref{eq:label_smoothed_cross_entropy_C_Core} first reduces the confidence placed on the oracle's labels by a small percentage $\epsilon$, and then encourages reliance on the C-Core attention vectors by reducing the additional loss term by a factor of $1-\epsilon$ whenever the predicted class by the model agrees with the C-Core attention vector based assignment. Assuming a perfect model again, in this case at noise rate $r=0$ the first term in the loss $L \approx 0$, while the second term is always $\epsilon\log p(\hat{y}=\hat{y}_{cc}|x_{i})$ due to the indicator variable $I$. Looking at this differently, the smoothed loss uses both the oracle's labels and the C-Core vectors as the ground. As $r \to 1$, a larger proportion of the oracle's labels is incorrect, meaning the first term of the loss will explode to infinity as is the case with cross-entropy, while the second term is independent of the noise rate and heavily contributes to the loss for deviations from the C-Core attention vectors based assignment. This means at high label noise rates, the smoothed loss leads to weight updates with reduced influence from noisy labels and thus provides more robust learning. 

Now assuming a random model, for low noise rates $r = 0$, the oracle's labels $p(y|x_{i})$ are correct, and $p(\hat{y}|x_{i})$ is a vector with entries $\frac{1}{C}$. The cross-entropy loss $L \to -log \frac{1}{C}$, and as $r \to 1: L \to -log \frac{1}{C}$. The smoothed loss has the same properties as the cross-entropy loss in both low and high label noise rate settings when the model randomly assigns labels. In summary, if we start with a poorly performing model after training on the clean random set, we expect no performance gains in using a smoothed loss function in both low and high-label noise settings. On the other hand, starting with a strong model and thus representative C-Core attention vectors, we have shown that label smoothing reduces the adverse effects of significant label noise rates using our custom loss function. In the next section, we present the experimental setup, models trained, AL strategies tested, datasets used, and the training configurations.

\section{Experimental Setup}\label{setup}
In this section, we describe the experimental setup that forms a high-dimensional grid of different configurations in active learning for image classification in the presence of label noise. We vary several DL architectures (4), AL algorithms (6), and benchmark datasets (4), as well as the Oracle label noise rates (4). Two of the DL models used in this work are CNN-based while the other two are ViT-based. All CNN-based models are trained in an AL setting with 5 of the 6 AL strategies, and all datasets over all the 4 label noise rates. The 6th AL strategy is GCI-ViTAL and is unique to ViTs. The ViT models in this work are trained under all six AL strategies, all datasets, and label noise settings.

\subsection{Deep Learning Models}\label{subsec:deep_leaarning_models}
For all four DL architectures in this work, the weights are transferred from the pre-training of the model on the ImageNet-1k dataset \cite{Russakovsky:ILSVRC15}. We then fine-tune the fully connected layer for classification. The CNN-based models used in this work are: ResNet34 \cite{He:ResNetL16} and VGG19 \cite{Russakovsky:VGG14, Simonyan:VDCNN15}, chosen for their popularity in image classification benchmarks as well as good performance. The ViT-based models of choice in this work are the base ViT with 14 non-overlapping 16 by 16 patches and 12 attention heads, and the Swin transformer \cite{liu:swin21}, which implements overlapping shifted patches as opposed to a grid of rigid patches. In Table \ref{tab:model_comparison} we show the model sizes in megabytes, as well as the number of frozen and trainable parameters used in our experiments.

\begin{table}[htbp]
    \centering
        \begin{tabular}{c|c|c|c|c|c}
            \hline
            Model & Size (MB) & Trainable Parameters & Frozen Parameters & Trainable/Frozen & Feature Extractor \\ \hline \hline
            ResNet34 & 44.7 & 5,130 & 11.2M & \textbf{0.0458}\%  & CNN \\ 
            VGG19 & \textbf{548} & \textbf{40,970} & \textbf{139.6M} & 0.0294\%  & CNN \\
            ViT Base & 330 & 7,690 & 85.7M & 0.0089\%  & Transformer \\ 
            Swin Transformer & 336 & 10,250 & 86.91M & 0.0118\% & Transformer \\ \hline
        \end{tabular}
        \vspace{0.3cm}
        \caption{Comparison of ResNet34, VGG19, ViT Base, and Swin Transformer in terms of model size, trainable parameters, frozen parameters, and feature extractor. It is worth noting that the older VGG19 model is larger and carries more parameters than all the other newer models.}
        \label{tab:model_comparison}
\end{table}

\subsection{Active Learning Algorithms}\label{subsec:ActiveLearning}
 We compare the following standard DAL query strategies: random query, information entropy-based selection, margin sampling, hybrid uncertainty and diversity, model delta, and ours, GCI-ViTAL. Due to a large number of experiments as well as training time, we limit AL strategies to the above six. We briefly explain each method below:

\begin{itemize}
  \item \textbf{Random Query:} This is the simplest query strategy, while also relatively effective. This AL strategy simply selects candidate images for labeling with equal probability from the unlabeled dataset. It does not take into account any information about the unlabeled data except its size, so it is likely to select samples that are not very informative for the model. 
  
  \item \textbf{Information Entropy-Based Selection:} This query strategy selects samples with the highest information entropy, which is a measure of uncertainty. Information entropy is high when the model is not confident about the predicted class of a sample. Selecting samples with high information entropy can help the model learn class boundaries, hence improving its performance.

  \item \textbf{Margin Sampling:} This query strategy selects samples with the smallest margin between the predicted scores of the two most likely classes. The margin is a measure of the confidence of the model in its prediction. Selecting samples with small margins helps label samples that better define decision boundaries of very similar classes.
  
  \item \textbf{Hybrid Uncertainty and Diversity:} This query strategy selects samples in a way that balances uncertainty and diversity from the labeled data. The uncertainty is measured by the information entropy of the sample, and diversity is measured by the distance of the sample to the labeled data in pixel space.
  
  \item \textbf{Model Delta:} This query strategy simply selects samples that are most likely to change the model's predictions if they were labeled. The model delta is calculated by comparing the predicted scores of two consecutive model states, that is the previously trained model at time $t-1$ and the model at time $t$. Samples with the highest change in the class probability distribution are selected for labeling.  
  
  \item \textbf{GCI-ViTAL (Ours):} This query strategy uses the final transformer block's output to select samples that are hard to classify to the current model, and are semantically different from their supposed class average image. This potentially means the points are a special case, most likely close to a decision boundary, and thus worth obtaining a label for from the oracle. This strategy has the advantage of considering the semantical similarities of images while making sample selections, and not only relying on the trained model, which is prone to adverse effects due to label noise.
  
\end{itemize}
 
\subsection{Datasets}\label{subsec:datasets}
\begin{figure*}
  \centering
  \begin{subfigure}[b]{0.45\textwidth}
    \includegraphics[width=\textwidth]{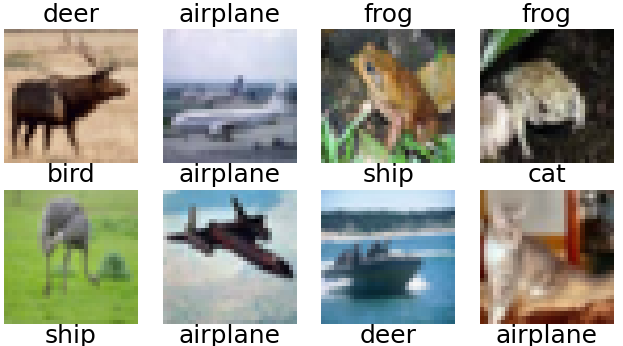}
    \caption{CIFAR10 random images}
  \end{subfigure}
   \rule{0.4pt}{150pt}
  \begin{subfigure}[b]{0.45\textwidth}
    \includegraphics[width=\textwidth]{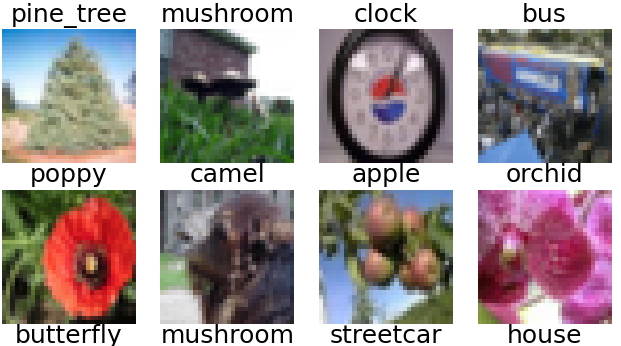}
    \caption{CIFAR100 random images}
  \end{subfigure}
  \caption{Random images from CIFAR-10 (first four columns) and CIFAR-100. Each image is of size 32 $\times$ 32 in the dataset.}
  \label{fig:CIFAR_10_100}
\end{figure*}

\begin{figure*}
  \centering
  \begin{subfigure}[t]{0.45\textwidth}
    \includegraphics[width=\textwidth]{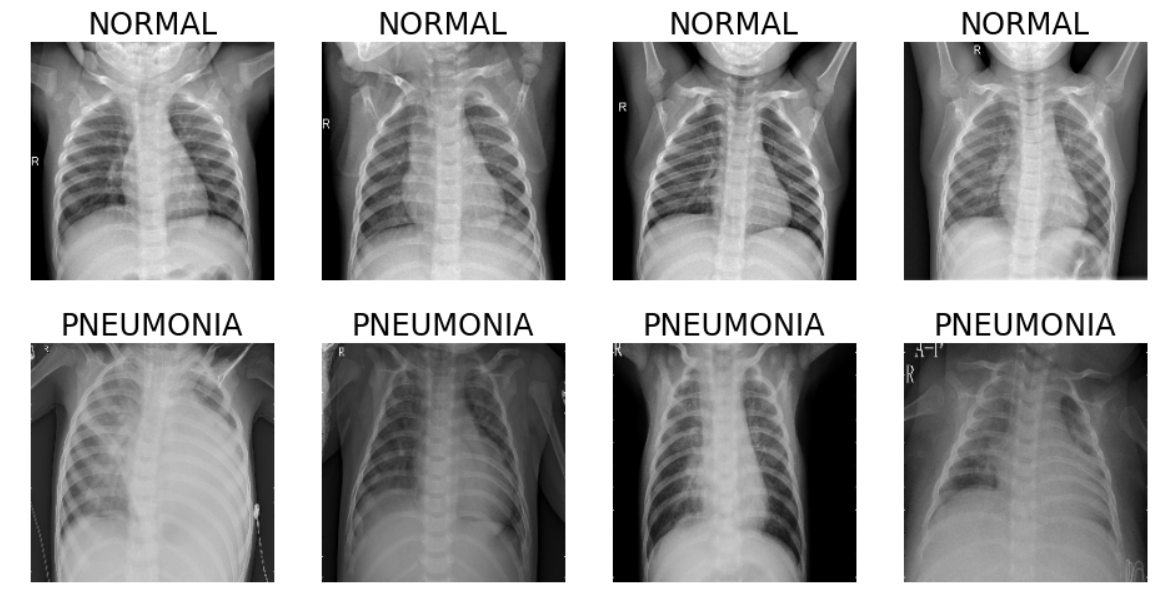}
    \caption{Chest X-ray images showing four pneumonic and normal scans}
  \end{subfigure}
   \rule{0.4pt}{150pt}
  \begin{subfigure}[t]{0.45\textwidth}
    \includegraphics[width=\textwidth]{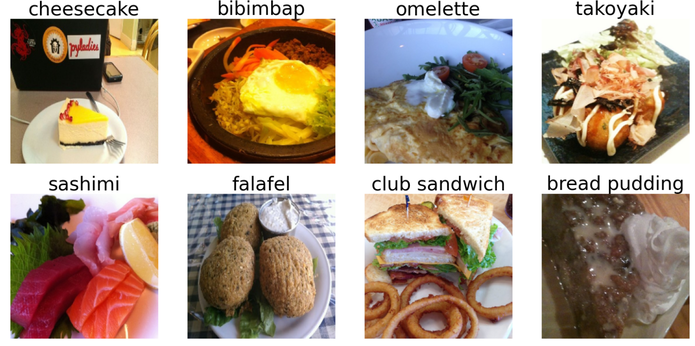}
    \caption{Food101 sample images with their class labels}
  \end{subfigure}
  \caption{Random images from the Chest X-ray of size 224 $\times$ 224 (first four columns), and the more complex Food101 dataset images of size 512 $\times$ 512}
  \label{fig:Chest_Xray_Food101}
\end{figure*}

 Due to the lack of benchmarks on DAL with label noise using multiple datasets for image classification, we run experiments on the following datasets: Chest X-ray Images (Pneumonia) \cite{kermany:chestXRay18}, Food101 \cite{bossard:Food10114}, CIFAR10 and  CIFAR100 \cite{Krizhevsky:CIFAR10009}. Table \ref{tab:datasets} below is a brief summary of each dataset. Figures \ref{fig:CIFAR_10_100} and \ref{fig:Chest_Xray_Food101} present sample image examples from each dataset.

\begin{table}[!htb]
  \centering
  \begin{tabular}{|p{1.9cm}|p{0.8cm}|p{1.6cm}|p{1.1cm}|p{1.6cm}|p{6.3cm}|}
    \hline
    Dataset & Year & \# Images & Classes & Image Sizes & Short description \\
    \hline
    \hline
    Chest X-ray (Pneumonia) & 2018 & 5,863 & 2 & $224 \times 224$  & Chest X-ray images used for pneumonia detection. \\
    \hline
    CIFAR10 & 2009 & 60,000 & 10& $32 \times 32$ & Small color images of objects from 10 common object classes. \\
    \hline
    CIFAR100 & 2009 & 60,000 & 100 & $32 \times 32$ & Extending CIFAR10, 100 classes grouped into 20 super-classes like vehicles, animals, and flowers, with finer sub-classes. \\
    \hline
    Food101 & 2004 & 101,000 & 101 & $512 \times 512$ & Images of food items categorized into 101 different classes. \\
    \hline
  \end{tabular}
  \vspace{0.3cm}
  \caption{Image classification datasets commonly used for deep active learning and the training of DL with noisy labels}
  \label{tab:datasets}
\end{table}

 For all the datasets, we use $66\%,17\%, 17\%$ train, validation, and test splits as pre-split in the CIFAR 10 and 100 datasets. All of these data splits are kept mutually exclusive to each other. The process for injecting label noise and simulating an oracle is as follows: First, we put aside a clean test set that is the same size as what is commonly used for each dataset. We also draw a clean random sample, without replacement, of 1024 images and their labels from the training set (independent of the number of classes), that will be used as an initial clean set for active learning as is customary. The training set labels are then corrupted with four levels of noise $r \in \{0,0.2,0.4,0.6\}$, by replacing the actual label of a sample with a randomly selected class, where each class has an equal probability of being selected. Once class-independent label noise has been injected, we use the number of training samples for validation that correspond to $17\%$ of the total dataset. On the active learning training scheme, no fixed labeling budget is set; we train and evaluate the models on the test set after each batch of AL selection and oracle labeling until the entire training set has been used. This has the advantage of allowing for analysis of model performance in both low-budget and high-budget active learning regimes in the context of a noisy oracle. This means we never stop the AL algorithm based on the labeling budget in developing the best algorithm. We, however, monitor test performance across labeling budget ticks to be able to compare models and AL query strategies for different datasets and noise labels on varying budgets. 
 
\subsection{Training Configuration}\label{subsec:Training_configuration}
While the training procedure is straightforward in a theoretical framework, the implementation details contain non-trivial components. For one, training in TensorFlow allows for more control of the training loop and thus makes it easier to incorporate active learning and noise injection. However, this comes with a disadvantage when it comes to training efficiency on a GPU. While PyTorch makes training on a GPU simple with data loaders, the process complicates the active learning portion of the cycle as the indices that come out of the data loader do not directly correspond to the indices in the dataset as all operations take and spit out tensors of size $batch\_size$, each represented by one index value. We incur extra compute in remapping the indices to match the AL format. We make use of PyTorch's distributed data parallelism and train on two NVIDIA RTX A4000 and two RTX A5000 GPUs, each with approximately 16 Gigabytes of RAM, based on availability on one compute node. 

\begin{figure*}[!htbp]
  \centering
    \includegraphics[width=0.75\textwidth]{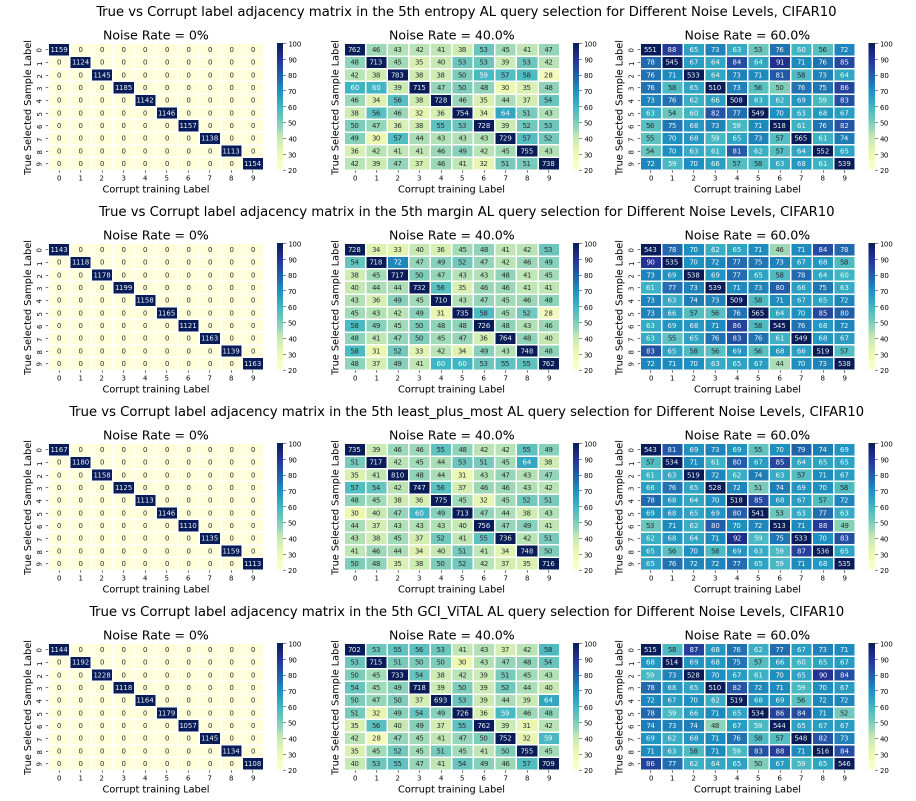}
    \caption{A depiction of the adjacency matrix of samples selected using four different AL strategies, their true training labels as well as the noisy training labels after noise injection up to 60\% label noise. The key takeaway from this view is that there is no statistical difference in the adjacency matrices of the selected samples due to label noise injection based on the AL strategy.}
    \label{figs:Noise_injection_per_AL_noise}
\end{figure*}

\begin{figure*}[!htbp]
  \centering
    \includegraphics[width=0.6\textwidth]{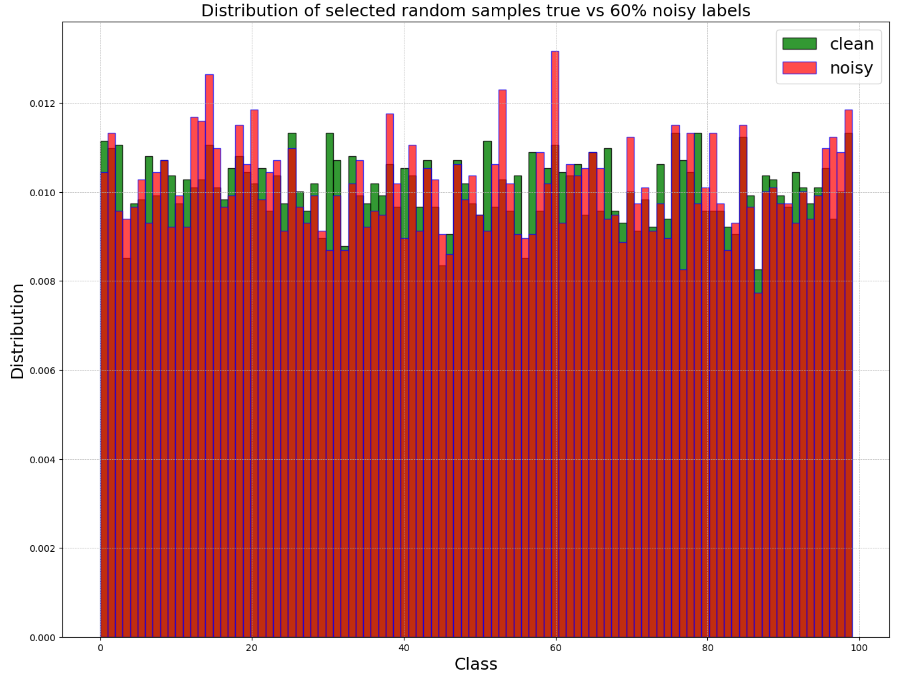}
    \caption{This shows an overlay of the distributions of the real and noisy labels at 60\% label noise for CIFAR100. The darker shade is where the two intersect or overlap. The selection of samples per AL query does not depend on the unknown real training labels for any of the AL strategies, neither does the label noise depend on the AL strategy, and so the picture looks similar for all AL strategies.}
  \label{figs:distribution_noisy_clean_60}
\end{figure*}

In training both CNN and ViT models, a learning rate scheduler is used, that reduces the learning rate by $10\%$ after 10 epochs of no improvement on the validation set. We also train with the early stopping of tolerance 5 epochs, meaning we halt training if the model hits a running 5 epoch validation loss maximum. This is a good indication that the model is beginning to overfit the training data. We select samples individually and not the best batch for AL algorithms that are not batch-based. We use the cross-entropy loss across all models except on the GCI-ViTAL AL strategy where we use a C-Core attention vector label smoothed cross-entropy loss. We use the Adam optimizer for ResNet and ViT. Stochastic gradient descent is the recommended optimizer for VGG19 based on the original paper and the choice in this work. The image transformations such as random cropping, image resizing, and pixel normalization we use for each model are those used by the authors of the respective models on the datasets we focus on in this work. This is sometimes necessary as the model architecture asserts a specific image size, such as is the case with base ViT, which expects images of size $224$. Images were resized to $32$ by $32$ for the CNN-based models and resized to $224$ by $224$ for the ViT models. For both CNN and ViT networks, training halts when all training samples are used, but we monitor the AL result after each labeling round.

Figures \ref{figs:Noise_injection_per_AL_noise} and \ref{figs:distribution_noisy_clean_60} present the adjacency matrix as well as real to corrupt label distribution in a given AL query selection cycle on CIFAR10 and CIFAR100 respectively. These figures demonstrate that we inject the same uniform label noise irrespective of the AL strategy, and thus the difference in performance will be driven mostly by the informativeness of the samples for most AL strategies, but the C-Core smoothed loss function as well in the case of GCI-ViTAL.

\FloatBarrier

\section{Results}\label{sec:results}
In this section, we present the test results of different DL models trained and tested on CIFAR10, 100, and the Chest X-ray (Pneumonia) datasets. We compare the results of multiple DAL algorithms including our proposed GCI-ViTAL, and show results for different label noise rates. We also compare the DAL algorithms on the test performance per labeling budget level. This helps show how the DAL algorithms compare in both low- and high-budget settings in the presence of label noise. Test performance against labeling budget plots can be found in appendix \ref{sec_res_of_results}

\begin{table}[!htbp]
\centering
\small
\begin{tabular}{@{}cccccccc@{}}
\toprule
\multirow{3}{*}{Noise Rate} & \multirow{2}{*}{Model} & \multicolumn{6}{c}{Active Learning Strategy} \\ \cmidrule(l){3-8} 
 &  & Random & Entropy & Margin & Hybrid & Model Delta & GCI-ViTAL\\ \cmidrule(lr){1-1} \cmidrule(l){2-8}
\multirow{3}{*}{0} & VGG19  & 0.8278 & 0.8288 & 0.8273 & 0.8370 & 0.8287 & -\\
 & ResNet34 & 0.8132 & 0.8319 & 0.8398 & 0.8348 & 0.8328 & -\\ 
 & ViTb16 & 0.9330 & 0.9320 & 0.9322 & 0.9338 & \textbf{0.9345} & 0.9338\\
 & SwinV2b16 & 0.8946 & 0.8922 & 0.8945 & 0.8936 & 0.8936 & 0.8939\\\cmidrule(lr){1-1} \cmidrule(l){2-8}
\multirow{3}{*}{0.2} & VGG19 & 0.7821 & 0.7564 & 0.7575 & 0.7433 & 0.7137 & -\\
 & ResNet34 & 0.7843 & 0.7496 & 0.7426 & 0.7499 & 0.7278 & -\\ 
 & ViTb16 & 0.9176 & 0.9174 & 0.9173 & 0.9117 & 0.9185 & \textbf{0.9286}\\
 & SwinV2b16 & 0.8828 & 0.8887 & 0.8837 & 0.8808 & 0.8779 &0.8830\\\cmidrule(lr){1-1} \cmidrule(l){2-8}
\multirow{3}{*}{0.4} & VGG19 & 0.7406 & 0.7280 & 0.6646 & 0.6651 & 0.6296 & -\\
 & ResNet34 & 0.7485 & 0.6691 & 0.6968 & 0.6496 & 0.6009 & -\\ 
 & ViTb16 & 0.9093 & 0.9020 & 0.9047 & 0.9114 & 0.9127 & \textbf{0.9152}\\
 & SwinV2b16 & 0.8757 & 0.8789 & 0.8841 & 0.8791 & 0.8829 & 0.8874\\\cmidrule(lr){1-1} \cmidrule(l){2-8}
\multirow{3}{*}{0.6} & VGG19 & 0.7063 & 0.6450 & 0.6737 & 0.6612 & 0.4912 & -\\
 & ResNet34 & 0.7041 & 0.6234 & 0.6592 & 0.5575 & 0.4477 & -\\
 & ViTb16 & 0.9011 & 0.8677 & 0.8922 & 0.8933 & 0.8815 & \textbf{0.9117}\\ 
 & SwinV2b16 & 0.8777 & 0.8643 & 0.8738 & 0.8697 & 0.8670 & 0.8850\\ \bottomrule
\end{tabular}
\vspace{0.3cm}
\caption{Classification test accuracy after 40,000 training examples for ResNet34, VGG19, ViTb16, and SwinV2b16 on CIFAR10 using six active learning strategies and four label noise rates.}
\label{tab:test_accuracy}
\end{table}

\begin{table}[!htbp]
\centering
\small
\begin{tabular}{@{}cccccccc@{}}
\toprule
\multirow{3}{*}{Noise Rate} & \multirow{2}{*}{Model} & \multicolumn{6}{c}{Active Learning Strategy} \\ \cmidrule(l){3-8} 
 &  & Random & Entropy & Margin & Hybrid & Model Delta & GCI-ViTAL \\ \cmidrule(lr){1-1} \cmidrule(l){2-8}
\multirow{3}{*}{0} & ViTb16  & 0.7062 & 0.7053 & 0.7004 & 0.6996 & 0.7055 & 0.7022\\
 & SwinV2b16  & \textbf{0.7329} & 0.7327 & 0.7299 & 0.7325 & 0.7326 & 0.7312\\\cmidrule(lr){1-1} \cmidrule(l){2-8}\cmidrule(lr){1-1} \cmidrule(l){2-8}
\multirow{3}{*}{0.2} & ViTb16  & 0.7038 & 0.7035 & 0.7046 & 0.6983 & 0.7035 & 0.7025\\
 & SwinV2b16  & 0.7330 & 0.7286 & 0.7331 & 0.7302 & \textbf{0.7332} & 0.7324\\\cmidrule(lr){1-1} \cmidrule(l){2-8}\cmidrule(lr){1-1} \cmidrule(l){2-8}
\multirow{3}{*}{0.4} & ViTb16  & 0.7034 & 0.6993 & 0.7006 & 0.7006 & 0.6998  & 0.7082\\
 & SwinV2b16  & 0.7326 & 0.7304 & 0.7305 & 0.7325 & 0.7324 & \textbf{0.7339}\\\cmidrule(lr){1-1} \cmidrule(l){2-8}\cmidrule(lr){1-1} \cmidrule(l){2-8}
\multirow{3}{*}{0.6}& ViTb16  & 0.7066 & 0.7055 & 0.7020 & 0.6981 & 0.7017 & 0.7023\\ 
 & SwinV2b16  & 0.7320 & 0.7293 & 0.7317 & 0.7307 & 0.7312 & \textbf{0.7326}\\
 \bottomrule
\end{tabular}
\vspace{0.3cm}
\caption{Classification test accuracy after 73,000 training examples for ViTb16 and SwinV2b16 on the Food101 dataset using six active learning strategies and four label noise rates. We do not run experiments for CNN models on this dataset due to resource and time constraints}
\label{tab:test_accuracy_food101}
\end{table}

\begin{table}[!htbp]
\centering
\small
\begin{tabular}{@{}cccccccc@{}}
\toprule
\multirow{3}{*}{Noise Rate} & \multirow{2}{*}{Model} & \multicolumn{6}{c}{Active Learning Strategy} \\ \cmidrule(l){3-8} 
 &  & Random & Entropy & Margin & Hybrid & Model Delta & GCI-ViTAL \\ \cmidrule(lr){1-1} \cmidrule(l){2-8}
\multirow{3}{*}{0} & VGG19  & 0.5691 & 0.5657 & 0.5720 & 0.5675 & 0.5605 & -\\
 & ResNet34 & 0.5698 & 0.5621 & 0.5657 & 0.5646 & 0.5637 & -\\
 & ViTb16 & 0.7327& 0.7359 & 0.7357 & 0.7347 & 0.7339 & \textbf{0.7504}\\
 & SwinV2b16 & 0.7146 & 0.7168 & 0.7106 & 0.7130 & 0.7130 & 0.7349\\\cmidrule(lr){1-1} \cmidrule(l){2-8}\cmidrule(lr){1-1} \cmidrule(l){2-8}
\multirow{3}{*}{0.2} & VGG19 & 0.5116 & 0.4605 & 0.4588 & 0.4612 & 0.4495 & -\\
 & ResNet34 & 0.5135 & 0.4582 & 0.4711 & 0.4867 & 0.4499 & -\\ 
 & ViTb16 & 0.7316 & 0.7287 & 0.7316 & 0.7287 & \textbf{0.7341} & 0.7318\\
 & SwinV2b16 & 0.6973 & 0.6978 & 0.6964 & 0.6951 & 0.6986 & 0.7064\\\cmidrule(lr){1-1} \cmidrule(l){2-8}\cmidrule(lr){1-1} \cmidrule(l){2-8}
\multirow{3}{*}{0.4} & VGG19 & 0.4595 & 0.3911 & 0.3899 & 0.3719 & 0.3400 & -\\
 & ResNet34 & 0.4664 & 0.3795 & 0.3679 & 0.3833 & 0.3359 & -\\
 & ViTb16 & 0.7194 & 0.7180 & 0.7146 & 0.7160 & 0.7177 & \textbf{0.7325}\\
 & SwinV2b16 & 0.6771 & 0.6786 & 0.6698 & 0.6749 & 0.6767 & 0.6800\\\cmidrule(lr){1-1} \cmidrule(l){2-8}\cmidrule(lr){1-1} \cmidrule(l){2-8}
\multirow{3}{*}{0.6} & VGG19 & 0.4084 & 0.3009 & 0.2930 & 0.3246 & 0.2204 & -\\
 & ResNet34 & 0.3946 & 0.3097 & 0.3196 & 0.3639 & 0.2210  & -\\ 
 & ViTb16 & 0.7075 & 0.6989 & 0.6873 & 0.6936 & 0.6959 & \textbf{0.7215}\\ 
 & SwinV2b16 & 0.6468 & 0.6426 & 0.6403 & 0.6389 & 0.6361 & 0.6441\\ \bottomrule
\end{tabular}
\vspace{0.3cm}
\caption{Classification test accuracy after 40,000 training examples for ResNet34, VGG19, ViTb16 and SwinV2b16 on CIFAR100 using six active learning strategies and four label noise rates. GCI-ViTAL outperforms the rest in high label noise rates due to its ability to handle label noise. CNN-based models deteriorate considerably compared to ViTs for increasing class count under label noise.}
\label{tab:test_accuracy_cifar100}
\end{table}

\begin{table}[!htbp]
\centering
\small
\begin{tabular}{@{}cccccccc@{}}
\toprule
\multirow{3}{*}{Noise Rate} & \multirow{2}{*}{Model} & \multicolumn{6}{c}{Active Learning Strategy} \\ \cmidrule(l){3-8} 
 &  & Random & Entropy & Margin & Hybrid & Model Delta & GCI-ViTAL \\ \cmidrule(lr){1-1} \cmidrule(l){2-8}
\multirow{3}{*}{0} & VGG19  & 0.6250 & 0.3189 & 0.4183 & 0.2724 & 0.5128 & - \\
 & ResNet  & 0.4359 & 0.2115 & 0.2115 & 0.3237 & 0.6138 & -\\
 & ViTb16  & 0.6250 & 0.5865 & 0.5737 & 0.5833 & 0.6089 & 0.6109\\
 & SwinV2b16  & \textbf{0.8798} & 0.8413 & 0.8365 & 0.8654 & 0.8589 & 0.8620\\\cmidrule(lr){1-1} \cmidrule(l){2-8}\cmidrule(lr){1-1} \cmidrule(l){2-8}
\multirow{3}{*}{0.2} & VGG19  & 0.4359 & 0.2772 & 0.1778 & 0.3589 & 0.5513 & -\\
 & ResNet  & 0.5321 & 0.4070 & 0.3429 & 0.3605 & 0.6250 & -\\ 
 & ViTb16  & 0.6298 & 0.5801 & 0.5993 & 0.6186 & 0.6089 & 0.6275\\
 & SwinV2b16  & 0.8782 & 0.8381 & 0.8542 & 0.8574 & 0.8685 & \textbf{0.8807}\\\cmidrule(lr){1-1} \cmidrule(l){2-8}\cmidrule(lr){1-1} \cmidrule(l){2-8}
\multirow{3}{*}{0.4} & VGG19  & 0.4439 & 0.2051 & 0.5352 & 0.1939 & 0.2901 & -\\
 & ResNet  & 0.5465 & 0.3701 & 0.6266 & 0.4535 & 0.6233 & -\\
 & ViTb16  & 0.6234 & 0.6201 & 0.6250 & 0.6346 & 0.6282  & 0.6368\\
 & SwinV2b16  & \textbf{0.8558} & 0.8349 & 0.8509 & 0.8477 & 0.8285 & 0.8510\\\cmidrule(lr){1-1} \cmidrule(l){2-8}\cmidrule(lr){1-1} \cmidrule(l){2-8}
\multirow{3}{*}{0.6}& VGG19  & 0.5208 & 0.3766 & 0.5144 & 0.5496 & 0.4936 & -\\
 & ResNet  & 0.3878 & 0.3766 & 0.2212 & 0.3253 & 0.4262 & -\\ 
 & ViTb16  & 0.5946 & 0.6362 & 0.5769 & 0.6201 & 0.6169 & 0.6273\\ 
 & SwinV2b16  & \textbf{0.8814} & 0.8654 & 0.7468 & 0.8381 & 0.8494 & 0.8661\\
 \bottomrule
\end{tabular}
\vspace{0.3cm}
\caption{Classification test accuracy after 40,000 training examples for ResNet34, VGG19, ViTb16 and SwinV2b16 on the Chest X-ray dataset using six active learning strategies and four label noise rates. Random selection using Swin transformer performs best in this binary classification case since the impact of label noise is not as announced as it is in the case of 10 or 100 classes.}
\label{tab:test_accuracy_chest_xray}
\end{table}

\begin{itemize}
    \item The first finding in our study that has support in the literature for non-active learning image classification is the superiority of ViTs over CNN-based models in label noise DAL. We observe that for three of the datasets in Tables \ref{tab:test_accuracy}, \ref{tab:test_accuracy_cifar100}, and \ref{tab:test_accuracy_chest_xray} in this study, transformer-based models (ViTb16 and SwinV2b16) performed significantly better than CNN-based models (ResNet34 and VGG19) out of the box, across all AL strategies and label noise rates. The performance difference is more pronounced in CIFAR100. CIFAR100 has 10 times more class labels while maintaining the same number of training examples as CIFAR10, while the Chest X-ray dataset presents complexity in the input image since the images are of much higher resolution than CIFAR10 as seen in Table \ref{tab:datasets}, and require expert opinion to tell positive and negative classes apart. There is very little difference in performance of the DAL methods on the Food101 ddataset across label noise rates.
    
    \item With regards to the DAL strategies across datasets, it is worth noting that, contrary to most literature, and in direct support of the findings in the following works \cite{Ren:DALSurvey20,Althammer:RandomSelectionComparision23,Mots'oehli:DeepActiveLabelNoise23}, we find that random query selection, while simple to implement and computationally efficient, posts comparable or superior classification accuracy results in DAL with label noise compared to margin, entropy, hybrid, and model delta based selection when out-of-the-box DL models are used without extensive model hyper-parameter optimization. We find this to be especially true for higher label noise rates. Our results show that sample selection that is only guided by a model trained on partially noisy labels selects non-optimal samples compared to random selection. This is not the case for GCI-ViTAL since the C-Core attention component of the AL query strategy does not get updated with noisy information once trained on the clean initial subset of the data, leading to noise resilience.
    
    \item Model delta, unlike all the other AL strategies in this study, selects samples based on the delta confidence of the previous model and the most recent model update. Samples with a large model confidence change are selected for labeling. We find that model delta AL performs better than entropy, margin, and hybrid in no-label noise settings, and tends to perform worse in increasing label noise settings. The empirical results in the high label noise setting make intuitive sense in that the introduction of label noise during training leads to a discrepancy between the predicted probabilities of consecutive models especially earlier in the training so that there could be model collapse. This behavior only persists in CNN-based models, i.e. model delta performs on par with all other AL strategies for zero label noise, but up to 45\% worse at 60\% label noise rate. Model delta however performs on par with other baseline AL strategies even for high label noise rates when the learner is a transformer-based model. This sheds more light on the robustness and high calibration of transformer-based models.
    
    \item GCI-ViTAL's higher generalization performance at high label noise rates comes at a higher computational cost in attaining the samples to be labeled as well as computing its custom loss function that is based on the C-Core attention assignment for each sample. We found that GCI-ViTAL runs as slow as model delta and approximately 2 times slower than entropy-based selection, margin, and hybrid uncertainty \& diversity selection. The difference in total runtime increases in the case of GCI-ViTAL since for each new retrained model, the attention maps are computed over the entire unlabeled dataset and stored for computing the Frobenious norm to the C-Core vectors and selecting the next training samples until the labeling budget is exhausted. Table \ref{tab:runtimes} shows the average runtime for each DAL algorithm per AL cycle. The random query strategy has the fastest runtime as expected since it does not depend on the input images nor the predicted probabilities from the currently trained model. However, we see a benefit in using GCi-ViTAL in the high-label noise regime with a large label budget.

    \item We empirically observe that GCI-ViTAL's robustness under label noise is more pronounced with the increasing number of classes in the classification problem. On CIFAR100 and Food101, ViT with GCI-ViTAL performs equal or marginally better than all other AL strategies under 40\% and 60\% label noise as can be seen in Tables \ref{tab:test_accuracy_cifar100}, \ref{tab:test_accuracy_food101} and Figures \ref{fig:vit_cifar100_results}, \ref{fig:swin_cifar100_results}, \ref{fig:swin_food101_results}, and \ref{fig:vit_food101_results} as opposed to performance on CIFAR10 with 10 classes shown in Figures \ref{fig:vit_cifar10_results} and \ref{fig:swin_cifar10_results}, as well as in the Chest X-ray dataset with just 2 classes shown in Figures \ref{fig:vit_chest_xray_results}, and \ref{fig:swin_chest_xray_results}. See Appendix \ref{sec_res_of_results} for all test accuracy plots.
    
    We hypothesize this is due to label smoothing guided by the C-Core Frobenious norm in the following manner: As the label noise rate increases up to 60\% in a 100-class problem, it means each sample selected for labeling is 60\% likely to be labeled erroneously as one of the other 99 classes. This leads to incorrect weight updates and thus large training errors. However, with GCI-ViTAL, the loss function is partially grounded in the small clean label distribution to redistribute the oracle's labeling in such a manner it agrees with the pre-trained model's semantical assignment of images to classes based on the learned representations from clean pre-training. Recall from Section \ref{subsub:GCI-VITAL} that while the loss is designed to tackle label noise, the C-Core assignment is not affected by label noise as we only train the MLP layer of the ViT on noisy labels, and not the feature extraction multi-head self-attention layers.
\end{itemize}


\begin{table}[htbp]
\centering
\begin{tabular}{l|c}
\hline
\textbf{Algorithms} & \textbf{Average Run Time (seconds) per AL Cycle} \\ \hline \hline
Random Query & 0.593 \\
Margin Sampling & 87.42 \\
Entropy-Based Selection & 87.33 \\
Hybrid Uncertainty and Diversity & 87.02 \\
Model Delta & 164.3 \\ 
GCI-ViTAL (ours) & 163.6\\ \hline
\end{tabular}
\vspace{0.31cm}
\caption{The average time it takes to select samples using different AL strategies on CIFAR10. It does not include the time required to train the model. However, for AL strategies that select samples based on inference on the unlabeled dataset such as Entropy-Based Selection, Hybrid Uncertainty and Diversity, and GCI-ViTAL, the inference time and time spent on the ranking of unlabeled samples is included in the selection runtime per AL cycle.}
\label{tab:runtimes}
\end{table}

\newpage

\section{Discussions}\label{sec:discussion}
We show that GCI-ViTAL performs equal or marginally better than most of the AL strategies for lower label noise rates, statistically on par with random selection on low label noise rates, and observably better on higher label noise rates, particularly on CIFAR100 where 60\% label noise rate means any sample has a 60\% chance of being mislabeled as one of the other 99 classes. We attribute this to GCI-ViTAL possessing an in-build pseudo-memory bank of high-accuracy mappings from the C-Core attention maps to labels that come in handy when the oracle's labels are very noisy. The C-Core attention vectors help in selecting outlier samples for a supposed class, and at the same time discount the confidence in the oracle's label under high label noise if it does not match that of the C-Core Frobenius norm-based assignment. In this case, we can show that the additional use of the attention maps learned during clean label pre-training on ImageNet through multi-head attention is advantageous over random selection under high label noise rates. We do this without heavily searching the hyper-parameter space of the underlying DL model in such a manner that would benefit our AL strategy over baseline models, thus reflecting a fairer comparison under label noise.

As the label noise rate increases from 0\% up to 60\%, the test accuracy for all models and AL strategies declines considerably for CIFAR10 and CIFAR100, and only marginally for the Chest X-ray and Food101 datasets. Still, the decline in performance is steeper for the CNN-based models as compared to their transformer-based counterparts. One reason for this could be that the pre-training in the transformer captures adequate local dependencies while at the same time learning rich global dependencies between pixels so that fine-tuning on noisy labels has a less detrimental effect on the overall probability outputs of the model. This is not the case with CNN-based models that have an inductive bias to prioritize local dependencies in explaining the differences in class labels, thus making the transformer more robust to label noise than CNN-based models \cite{khanal:investigatingVitRobustness24,Mathilde:EmergingPropsViT21,Hendrycks:SSLforRobustness19}. To contrast the two in the face of high label noise, CNNs learn a smoother decision boundary between classes while transformers learn a more complex decision boundary \cite{Kolesnikov:ViT21,Ramachandran:Standaloneselfattention19,Cordonnier:OnTR2019}. This is because ViTs are larger models with considerably more parameters than CNNs as shown in Table \ref{tab:model_comparison}, and more parameters allow for a more fine-grained complex decision boundary between classes. The more parameters of transformers allow for the decision boundary to change due to label noise but only change so insignificantly that the change does not lead to the misclassification of many neighboring samples. However we note that the ViT is more robust to label noise in most cases that require complex decision boundaries since it has a smaller trainable-to-frozen parameters ratio, meaning it is forced to condense a lot more of the signal into useful weights, thus it is less likely to overfit to noisy training data as compared to CNN-based models. This leads us to the idea of aiming for larger frozen parameter models for feature extraction and low trainable parameters as future work.

 While the use of ViTs in active learning with label noise shows promising results, the high computational cost is a disadvantage that needs to be addressed. Possible ways of addressing this include deploying ViTs with fewer layers or reducing model size through methods such as quantization \cite{Rok:QuantizationSurvey23, Feng:PostTrainingQuantizationViTs23} and model distillation \cite{Hinton:DistillingTK15,Yang:ViTKDPG22,Das:SelfKDWithLabelNoise23}. We are also interested in exploring how ViT models of different sizes (small, base, and large) are impacted by label noise. This can be especially important in navigating the tradeoffs in performance and computational complexity. The more fundamental question that is a result of this work is: in the context of high label noise, at what point do the active learning and few-shot learning paradigms converge, what are the factors, and what role can weak or self-supervised learning play in improving generalization and robustness. Last but not least, in the wake of the advances in large language models (LLMs) as well as multi-modal learning, how can we leverage LLMs to produce text-guided active learning strategies, guide the oracle through caption generations, and create explainable ViT applications in the label noise domain? These ideas are partially inspired by the following works \cite{Hu:CrossModalNoisyLabels21,Wazzan:LLMBasedImageSearch24,chen:MultilingualImagePaLI23}.

 \subsection{Limitations}\label{subsec:limitations}
 We note that while this work addresses the use of out-of-the-box DL models to give a good reflection of realistic AL baselines, it lacks in providing a comprehensive comparison of the proposed method (GCI-ViTAL) to state-of-the-art AL methods with label noise. The work also does not address or show results of GCI-ViTAL on more complex real-world datasets outside the datasets covered due to time and resource constraints.

\section{Conclusion}\label{sec:conclusion}
In conclusion, this study provides valuable insights into the performance of various deep learning models trained and tested on CIFAR10, CIFAR100, Food101, and the Chest X-ray (Pneumonia) datasets, particularly in the context of active learning with label noise. We found that transformer-based models, such as ViT and Swin Transformer, consistently outperformed CNN-based models across all datasets and AL strategies, indicating their superiority in handling complex image classification tasks.  We attribute ViT's robustness to their ability to capture both local and global dependencies, resulting in more complex and elastic decision boundaries that are less affected by incorrectly labeled training samples. Furthermore, our results challenge conventional wisdom by showing that random query selection often yields superior classification accuracy in AL with label noise compared to more complex AL strategies that rely solely on uncertainty and diversity metrics. Despite these findings, it is worth noting that transformer-based models incur higher computational costs compared to CNN-based models. GCI-ViTAL, our proposed AL strategy, exhibits higher generalization performance at high label noise rates, albeit with increased computational cost. This strategy leverages the inherent robustness of transformer-based models by making use of the semantic relationships of images without an oracle's labels.

This work opens up several avenues for future research. Exploring methods to mitigate the computational overhead of transformer-based models, such as deploying smaller ViT models or applying model quantization and distillation techniques, could be a promising direction. Additionally, investigating the convergence of active learning and few-shot learning paradigms in the context of high label noise, as well as exploring the role of weak or self-supervised learning to improve generalization in label noise scenarios could provide further insights into enhancing model performance. We are also interested in investigating how ViT model size impacts DAL under label noise. In summary, our study contributes to the current understanding of DL models' behavior under DAL scenarios with label noise and suggests potential avenues for addressing challenges in this domain.

\section*{Acknowledgement}
We acknowledge technical support and computing resources from The University of Hawaii Information Technology Services – Cyberinfrastructure funded in part by the National Science Foundation CC* award \texttt{\#}2201428.

\newpage

\bibliographystyle{unsrt}  
\bibliography{templateArxiv} 

\begin{thebibliography}{10}

\bibitem{Ren:DALSurvey20}
P.~Ren, Y.~Xiao, X.~Chang, P.~Huang, Z.~Li, X.~Chen, and X.~Wang.
\newblock A survey of deep active learning.
\newblock {\em ACM Computing Surveys (CSUR)}, 54:1 -- 40, 2020.

\bibitem{Algan:ImageNoisySurvey}
G.~Algan and I.~Ulusoy.
\newblock Image classification with deep learning in the presence of noisy
  labels: A survey.
\newblock {\em Knowledge-Based Systems}, 215:106771, 2021.

\bibitem{Mots'oehli:DeepActiveLabelNoise23}
M.~Mots'oehli and K.~Baek.
\newblock Deep active learning in the presence of label noise: A survey.
\newblock {\em arXiv preprint arXiv:2302.11075}, 2023.

\bibitem{Zhang:NoiseMemorization16}
C.~Zhang, S.~Bengio, M.~Hardt, B.~Recht, and O.~Vinyals.
\newblock Understanding deep learning requires rethinking generalization.
\newblock {\em Communications of the ACM}, 64, 11 2016.

\bibitem{Malach:Decoupling17}
E.~Malach and S.~Shalev-Shwartz.
\newblock Decoupling "when to update" from "how to update".
\newblock In {\em Proceedings of the 31st International Conference on Neural
  Information Processing Systems}, NIPS'17, page 961–971, Red Hook, NY, USA,
  2017. Curran Associates Inc.

\bibitem{Han:Coteaching18}
B.~Han, Q.~Yao, X.~Yu, G.~Niu, M.~Xu, W.~Hu, I.~Tsang, and M.~Sugiyama.
\newblock Co-teaching: Robust training of deep neural networks with extremely
  noisy labels.
\newblock In {\em Advances in Neural Information Processing Systems}, 2018.

\bibitem{Liu:EarlyLearningMemorization20}
S.~Liu, J.~Niles-Weed, N.~Razavian, and C.~Fernandez-Granda.
\newblock Early-learning regularization prevents memorization of noisy labels.
\newblock In {\em Proceedings of the 34th International Conference on Neural
  Information Processing Systems}, NIPS'20, 2020.

\bibitem{Liang:NoisySurvey22}
X.~Liang, X.~Liu, and L.~Yao.
\newblock Review–a survey of learning from noisy labels.
\newblock {\em ECS Sensors Plus}, 1(2):021401, 06 2022.

\bibitem{Kolesnikov:ViT21}
A.~Kolesnikov, A.~Dosovitskiy, D.~Weissenborn, G.~Heigold, J.~Uszkoreit,
  L.~Beyer, M.~Minderer, M.~Dehghani, N.~Houlsby, S.~Gelly, T.~Unterthiner, and
  X.~Zhai.
\newblock An image is worth 16x16 words: Transformers for image recognition at
  scale.
\newblock In {\em 9th International Conference on Learning Representations,
  {ICLR} 2021}, 2021.

\bibitem{Touvron:GoingDW21}
H.~Touvron, M.~Cord, A.~Sablayrolles, G.~Synnaeve, and H.~J'egou.
\newblock Going deeper with image transformers.
\newblock {\em 2021 IEEE/CVF International Conference on Computer Vision
  (ICCV)}, pages 32--42, 2021.

\bibitem{Ridnik:ImageNer21ForThemasses21}
T.~Ridnik Tal, E.~Ben-Baruch, A.~Noy, and L.~Zelnik.
\newblock Imagenet-21k pretraining for the masses.
\newblock In J.~Vanschoren and S.~Yeung, editors, {\em Proceedings of the
  Neural Information Processing Systems Track on Datasets and Benchmarks},
  volume~1. Curran, 2021.

\bibitem{Yuan:IncorporatingCD21}
K.~Yuan, S.~Guo, Z.~Liu, A.~Zhou, F.~Yu, and W.~Wu.
\newblock Incorporating convolution designs into visual transformers.
\newblock {\em 2021 IEEE/CVF International Conference on Computer Vision
  (ICCV)}, pages 559--568, 2021.

\bibitem{Grriz:CostEffectiveA17}
M.~G{\'o}rriz, A.~Carlier, E.~Faure, and X.~Gir{\'o} i~Nieto.
\newblock Cost-effective active learning for melanoma segmentation.
\newblock {\em ArXiv}, abs/1711.09168, 2017.

\bibitem{Konyushkova:LearningAL17}
K.~Konyushkova, R.~Sznitman, and P.~Fua.
\newblock Learning active learning from data.
\newblock In {\em Proceedings of the 31st International Conference on Neural
  Information Processing Systems}, NIPS'17, page 4228–4238, Red Hook, NY,
  USA, 2017. Curran Associates Inc.

\bibitem{Kremer:BigUbi17}
J.~Kremer, K.~Stensbo-Smidt, F.~Gieseke, K.S. Pedersen, and C.~Igel.
\newblock Big universe, big data: Machine learning and image analysis for
  astronomy.
\newblock {\em IEEE Intelligent Systems}, 32(2):16--22, 2017.

\bibitem{Carena:NuclearInstruments14}
F.~Carena, W.~Carena, S.~Chapeland, V.~{Chibante Barroso}, F.~Costa, E.~Dénes,
  R.~Divià, U.~Fuchs, A.~Grigore, T.~Kiss, G.~Simonetti, C.~Soós, A.~Telesca,
  P.~{Vande Vyvre}, and B.~{von Haller}.
\newblock The alice data acquisition system.
\newblock {\em Nuclear Instruments and Methods in Physics Research Section A:
  Accelerators, Spectrometers, Detectors and Associated Equipment},
  741:130--162, 2014.

\bibitem{Gutleber:HighEnergy03}
J.~Gutleber, S.~Murray, and L.~Orsini.
\newblock Towards a homogeneous architecture for high-energy physics data
  acquisition systems.
\newblock {\em Computer Physics Communications}, 153(2):155--163, 2003.

\bibitem{Zhai:S4LSS19}
X.~Zhai, A.~Oliver, A.~Kolesnikov, and L.~Beyer.
\newblock S4l: Self-supervised semi-supervised learning.
\newblock {\em 2019 IEEE/CVF International Conference on Computer Vision
  (ICCV)}, pages 1476--1485, 2019.

\bibitem{Chen:ContrastiveVR20}
T.~Chen, S.~Kornblith, M.~Norouzi, and G.~Hinton.
\newblock A simple framework for contrastive learning of visual
  representations.
\newblock In {\em Proceedings of the 37th International Conference on Machine
  Learning}, ICML'20, pages 1597--1607. JMLR.org, 2020.

\bibitem{chen:SSViT21}
X.~Chen, S.~Xie, and K.~He.
\newblock An empirical study of training self-supervised vision transformers.
\newblock In {\em 2021 IEEE/CVF International Conference on Computer Vision
  (ICCV)}, pages 9620--9629, 2021.

\bibitem{He:MaskedAE22}
K.~He, X.~Chen, S.~Xie, Y.~Li, P.~Dollár, and R.~Girshick.
\newblock Masked autoencoders are scalable vision learners.
\newblock In {\em 2022 IEEE/CVF Conference on Computer Vision and Pattern
  Recognition (CVPR)}, pages 15979--15988, 2022.

\bibitem{assran:Ijepa23}
M.~Assran, Q.~Duval, I.~Misra, P.~Bojanowski, P.~Vincent, M.~Rabbat, Y.~LeCun,
  and N.~Ballas.
\newblock Self-supervised learning from images with a joint-embedding
  predictive architecture.
\newblock In {\em Proceedings of the IEEE/CVF Conference on Computer Vision and
  Pattern Recognition}, pages 15619--15629, 2023.

\bibitem{Lewis:PoolBased94}
D.~Lewis and W.~Gale.
\newblock A sequential algorithm for training text classifiers.
\newblock In {\em SIGIR '94}, pages 3--12, London, 1994. Springer London.

\bibitem{Cao:BALD21}
X.~Cao and I.~Tsang.
\newblock Bayesian active learning by disagreements: A geometric perspective.
\newblock {\em ArXiv}, abs/2105.02543, 2021.

\bibitem{Gal:BALD17}
Y.~Gal, R.~Islam, and Z.~Ghahramani.
\newblock Deep bayesian active learning with image data.
\newblock In {\em International Conference of Machine Learning}, volume
  abs/1703.02910, pages 1183--1192, 2017.

\bibitem{Sener:CoreSetActiveL18}
O.~Sener and S.~Savarese.
\newblock Active learning for convolutional neural networks: A core-set
  approach.
\newblock {\em International Conference on Learning Representations (Poster)},
  2018.

\bibitem{HarPeled:MaxMarginCoreSet07}
S.~Har-Peled, D.~Roth, and D.~Zimak.
\newblock Maximum margin coresets for active and noise tolerant learning.
\newblock In {\em International Joint Conferences on Artificial Intelligence},
  2007.

\bibitem{Younesian:ActiveLNoisyStrams20}
T.~Younesian, D.H. Epema, and L.~Chen.
\newblock Active learning for noisy data streams using weak and strong
  labelers.
\newblock {\em ArXiv}, abs/2010.14149, 2020.

\bibitem{Wei:NoisyAnnotation22}
J.~Wei, Z.~Zhu, H.~Cheng, T.~Liu, G.~Niu, and Y.~Liu.
\newblock Learning with noisy labels revisited: A study using real-world human
  annotations.
\newblock {\em 10th International Conference on Learning Representations},
  2022.

\bibitem{Chao:ThreeTeaching24}
G.~Chao, K.~Zhang, X.~Wang, and D.~Chu.
\newblock Three-teaching: A three-way decision framework to handle noisy
  labels.
\newblock {\em Applied Soft Computing}, 154:111400, 2024.

\bibitem{Gupta:NoisyBA20}
G.~Gupta, A.K. Sahu, and W.~Lin.
\newblock Noisy batch active learning with deterministic annealing.
\newblock In {\em arXiv: Learning}, 2020.

\bibitem{Amin:InducedAbstention21}
K.~Amin, G.~DeSalvo, and A.~Rostamizadeh.
\newblock Learning with labeling induced abstentions.
\newblock In M.~Ranzato, A.~Beygelzimer, Y.~Dauphin, P.S. Liang, and J.~Wortman
  Vaughan, editors, {\em Advances in Neural Information Processing Systems},
  volume~34, pages 12576--12586. Curran Associates, Inc., 2021.

\bibitem{Younesian:QActor21}
T.~Younesian, Z.~Zhao, A.~Ghiassi, R.~Birke, and L.~Chen.
\newblock Qactor: Active learning on noisy labels.
\newblock In Vineeth~N. Balasubramanian and Ivor Tsang, editors, {\em
  Proceedings of The 13th Asian Conference on Machine Learning}, volume 157 of
  {\em Proceedings of Machine Learning Research}, pages 548--563. PMLR, 17--19
  Nov 2021.

\bibitem{Huang:OracleEpiphany16}
T.~Huang, L.~Lihong, A.~Vartanian, S.~Amershi, and X.J. Zhu.
\newblock Active learning with oracle epiphany.
\newblock In D.~Lee, M.~Sugiyama, U.~Luxburg, I.~Guyon, and R.~Garnett,
  editors, {\em Advances in Neural Information Processing Systems}, volume~29.
  Curran Associates, Inc., 2016.

\bibitem{Yan:ImperfectLabelers16}
S.~Yan, K.~Chaudhuri, and T.~Javidi.
\newblock Active learning from imperfect labelers.
\newblock In {\em Proceedings of the 30th International Conference on Neural
  Information Processing Systems}, NIPS'16, page 2136–2144, Red Hook, NY,
  USA, 2016. Curran Associates Inc.

\bibitem{Roschewitz:ALDatasetQuality22}
M.~Roschewitz, D.~Coelho, R.~Tanno, A.~Schwaighofer, K.~Tezcan, M.~Monteiro,
  S.~Bannur, M.~Lungren, A.~Nori, B.~Glocker, J.~Alvarez-Valle, and O.~Oktay.
\newblock Active label cleaning for improved dataset quality under resource
  constraints.
\newblock {\em Nature Communications}, 13, 03 2022.

\bibitem{sheng2024foster}
Mengmeng Sheng, Zeren Sun, Tao Chen, Shuchao Pang, Yucheng Wang, and Yazhou
  Yao.
\newblock Foster adaptivity and balance in learning with noisy labels.
\newblock {\em arXiv preprint arXiv:2407.02778}, 2024.

\bibitem{li:empiricalEfficacy22}
Y.~Li, M.~Chen, Y.~Liu, D.~He, and Q.~Xu.
\newblock An empirical study on the efficacy of deep active learning for image
  classification.
\newblock 2022.

\bibitem{liu:swin21}
H.~Liu, G.~Cheng, W.~Lin, J.~Yang, J.~Yang, H.~Zhang, Z.~Zhang, and W.~Wu.
\newblock Swin transformer: Hierarchical vision transformer using shifted
  windows.
\newblock In {\em International Conference on Computer Vision (ICCV)}, 2021.

\bibitem{Han2021TransformerIT}
K.~Han, A.~Xiao, E.~Wu, J.~Guo, C.~Xu, and Y.~Wang.
\newblock Transformer in transformer.
\newblock In {\em Neural Information Processing Systems}, 2021.

\bibitem{Ding:DaViT22}
M.~Ding, B.~Xiao, N.~Codella, P.~Luo, J.~Wang, and L.~Yuan.
\newblock Davit: Dual attention vision transformers.
\newblock In {\em Computer Vision -- ECCV 2022}, pages 74--92, Cham, 2022.
  Springer Nature Switzerland.

\bibitem{fan:multiscale21}
H.~Fan, B.~Xiong, K.~Mangalam, Y.~Li, Z.~Yan, J.~Malik, and C.~Feichtenhofer.
\newblock Multiscale vision transformers.
\newblock In {\em Proceedings of the IEEE/CVF international conference on
  computer vision}, pages 6824--6835, 2021.

\bibitem{Caramalau:VisualTF2021}
R.~Caramalau, B.~Bhattarai, and T.~Kim.
\newblock Visual transformer for task-aware active learning.
\newblock {\em ArXiv}, abs/2106.03801, 2021.

\bibitem{HE:ViTMedical21}
H.~Kelei, G.~Chenand~L. Zhuoyuan, R.~Islem, Y.~Zihao, J.~Wen Ji, G.~Yang,
  W.~Qian, Z.~Junfeng, and S.~Dinggang.
\newblock Transformers in medical image analysis.
\newblock {\em Intelligent Medicine}, 3(1):59--78, 2023.

\bibitem{Rotman:MultiTaskALTransformerBased22}
G.~Rotman and R.~Reichart.
\newblock {Multi-task Active Learning for Pre-trained Transformer-based
  Models}.
\newblock {\em Transactions of the Association for Computational Linguistics},
  10:1209--1228, 11 2022.

\bibitem{Russakovsky:ILSVRC15}
O.~Russakovsky, J.~Deng, H.~Su, J.~Krause, S.~Satheesh, S.~Ma, Z.~Huang,
  A.~Karpathy, A.~Khosla, M.~Bernstein, A.C. Berg, and L.~Fei-Fei.
\newblock {ImageNet Large Scale Visual Recognition Challenge}.
\newblock {\em International Journal of Computer Vision (IJCV)},
  115(3):211--252, 2015.

\bibitem{Frobenius:UeberlineareSubstitutionen1877}
G.~Frobenius.
\newblock Ueber lineare substitutionen und bilineare formen.
\newblock {\em Journal für die reine und angewandte Mathematik}, 84:1--63,
  1877.

\bibitem{Kazakos:SpectralDM80}
D.~Kazakos and P.~Papantoni-Kazakos.
\newblock Spectral distance measures between gaussian processes.
\newblock In {\em IEEE International Conference on Acoustics, Speech, and
  Signal Processing}, 1980.

\bibitem{Kullback:KLdivergence51}
S.~Kullback and R.~A. Leibler.
\newblock {On Information and Sufficiency}.
\newblock {\em The Annals of Mathematical Statistics}, 22(1):79 -- 86, 1951.

\bibitem{Guo:2019AFN}
Pei-Chang Guo.
\newblock A frobenius norm regularization method for convolutional kernels to
  avoid unstable gradient problem.
\newblock {\em ArXiv}, abs/1907.11235, 2019.

\bibitem{Yuan:MaximumBatchFrobeniusNorm22}
Yuan Wu, Diana Inkpen, and Ahmed El-Roby.
\newblock Maximum batch frobenius norm for multi-domain text classification.
\newblock In {\em ICASSP 2022 - 2022 IEEE International Conference on
  Acoustics, Speech and Signal Processing (ICASSP)}, pages 3763--3767, 2022.

\bibitem{Qiu:ProbabilisticNormClustering18}
David Qiu, Anuran Makur, and Lizhong Zheng.
\newblock Probabilistic clustering using maximal matrix norm couplings.
\newblock In {\em 2018 56th Annual Allerton Conference on Communication,
  Control, and Computing (Allerton)}, page 1020–1027. IEEE Press, 2018.

\bibitem{spatialFrobeniousJaya:Ramnarayan19}
P.~Ramnarayan, S.~Subhashree, and B.~Pradyut.
\newblock Spectral clustering and spatial frobenius norm based jaya
  optimization for bandselection of hyperspectral images.
\newblock {\em IET Image Processing}, 13, 02 2019.

\bibitem{He:ResNetL16}
K.~He, X.~Zhang, S.~Ren, and J.~Sun.
\newblock Deep residual learning for image recognition.
\newblock In {\em IEEE Conference on Computer Vision and Pattern Recognition
  (CVPR)}, pages 770--778, 2016.

\bibitem{Russakovsky:VGG14}
O.~Russakovsky, J.~Deng, H.~Su, J.~Krause, S.~Satheesh, S.~Ma, Z.~Huang,
  A.~Karpathy, A.~Khosla, M.S. Bernstein, A.C. Berg, and L.~Fei{-}Fei.
\newblock Imagenet large scale visual recognition challenge.
\newblock In {\em CoRR}, volume abs/1409.0575, 2014.

\bibitem{Simonyan:VDCNN15}
K.~Simonyan and A.~Zisserman.
\newblock Very deep convolutional networks for large-scale image recognition.
\newblock {\em International Conference on Learning Representations},
  abs/1409.1556, 2015.

\bibitem{kermany:chestXRay18}
D.~Kermany, K.~Zhang, and M.~Goldbaum.
\newblock Labeled optical coherence tomography (oct) and chest x-ray images for
  classification.
\newblock {\em Mendeley data}, 2(2):651, 2018.

\bibitem{bossard:Food10114}
Lukas Bossard, Matthieu Guillaumin, and Luc Van~Gool.
\newblock Food-101 -- mining discriminative components with random forests.
\newblock In {\em European Conference on Computer Vision}, 2014.

\bibitem{Krizhevsky:CIFAR10009}
A.~Krizhevsky, V.~Nair, and G.~Hinton.
\newblock Learning multiple layers of features from tiny images.
\newblock {\em CIFAR-100 (Canadian Institute for Advanced Research)}, 2009.

\bibitem{Althammer:RandomSelectionComparision23}
S.~Althammer, G.~Zuccon, S.~Hofst\"{a}tter, S.~Verberne, and A.~Hanbury.
\newblock Annotating data for fine-tuning a neural ranker? current active
  learning strategies are not better than random selection.
\newblock In {\em Proceedings of the Annual International ACM SIGIR Conference
  on Research and Development in Information Retrieval in the Asia Pacific
  Region}, SIGIR-AP '23, page 139–149, New York, NY, USA, 2023. Association
  for Computing Machinery.

\bibitem{khanal:investigatingVitRobustness24}
B.~Khanal, P.~Shrestha, S.~Amgain, B.~Khanal, B.~Bhattarai, and C.A. Linte.
\newblock Investigating the robustness of vision transformers against label
  noise in medical image classification, 2024.

\bibitem{Mathilde:EmergingPropsViT21}
M.~Caron, H.~Touvron, I.~Misra, H.~Jegou, J.~Mairal, P.~Bojanowski, and
  A.~Joulin.
\newblock Emerging properties in self-supervised vision transformers.
\newblock In {\em 2021 IEEE/CVF International Conference on Computer Vision
  (ICCV)}, pages 9630--9640, 2021.

\bibitem{Hendrycks:SSLforRobustness19}
D.~Hendrycks, M.~Mazeika, S.~Kadavath, and D.~Song.
\newblock Using self-supervised learning can improve model robustness and
  uncertainty.
\newblock In {\em Proceedings of the 33rd International Conference on Neural
  Information Processing Systems}, Red Hook, NY, USA, 2019. Curran Associates
  Inc.

\bibitem{Ramachandran:Standaloneselfattention19}
P.~Ramachandran, N.~Parmar, A.~Vaswani, I.~Bello, A.~Levskaya, and J.~Shlens.
\newblock Stand-alone self-attention in vision models.
\newblock In {\em Proceedings of the 33rd International Conference on Neural
  Information Processing Systems}, 2019.

\bibitem{Cordonnier:OnTR2019}
J.~Cordonnier, A.~Loukas, and M.~Jaggi.
\newblock On the relationship between self-attention and convolutional layers.
\newblock {\em ArXiv}, abs/1911.03584, 2019.

\bibitem{Rok:QuantizationSurvey23}
B.~Rok, A.~Azarpeyvand, and A.~Khanteymoori.
\newblock A comprehensive survey on model quantization for deep neural networks
  in image classification.
\newblock {\em Association for Computing Machinery}, 14(6), nov 2023.

\bibitem{Feng:PostTrainingQuantizationViTs23}
K.~Feng, Z.~Chen, F.~Gao, Z.~Wang, L.~Xu, and W.~Lin.
\newblock Post-training quantization for vision transformer in transformed
  domain.
\newblock In {\em 2023 IEEE International Conference on Multimedia and Expo
  (ICME)}, pages 1457--1462, Los Alamitos, CA, USA, jul 2023. IEEE Computer
  Society.

\bibitem{Hinton:DistillingTK15}
G.E. Hinton, O.~Vinyals, and J.~Dean.
\newblock Distilling the knowledge in a neural network.
\newblock {\em ArXiv}, abs/1503.02531, 2015.

\bibitem{Yang:ViTKDPG22}
Z.~Yang, Z.~Li, A.~Zeng, Z.~Li, C.~Yuan, and Y.~Li.
\newblock Vitkd: Practical guidelines for vit feature knowledge distillation.
\newblock {\em ArXiv}, abs/2209.02432, 2022.

\bibitem{Das:SelfKDWithLabelNoise23}
R.~Das and S.~Sanghavi.
\newblock Understanding self-distillation in the presence of label noise.
\newblock In {\em Proceedings of the 40th International Conference on Machine
  Learning}, ICML'23. JMLR.org, 2023.

\bibitem{Hu:CrossModalNoisyLabels21}
P.~Hu, X.~Peng, H.~Zhu, L.~Zhen, and J.~Lin.
\newblock Learning cross-modal retrieval with noisy labels.
\newblock In {\em 2021 IEEE/CVF Conference on Computer Vision and Pattern
  Recognition (CVPR)}, pages 5399--5409, 2021.

\bibitem{Wazzan:LLMBasedImageSearch24}
A.~Wazzan, S.~MacNeil, and R.~Souvenir.
\newblock Comparing traditional and llm-based search for image geolocation.
\newblock In {\em Proceedings of the 2024 Conference on Human Information
  Interaction and Retrieval}, page 291–302, New York, NY, USA, 2024.
  Association for Computing Machinery.

\bibitem{chen:MultilingualImagePaLI23}
X.~Chen, X.~Wang, S.~Changpinyo, A.~Piergiovanni, P.~Padlewski, D.~Salz,
  S.~Goodman, A.~Grycner, B.~Mustafa, L.~Beyer, A.~Kolesnikov, J.~Puigcerver,
  N.~Ding, K.~Rong, H.~Akbari, G.~Mishra, L.~Xue, A.~Thapliyal, J.~Bradbury,
  W.~Kuo, M.~Seyedhosseini, C.~Jia, B.K. Ayan, C~Riquelme, A.~Steiner,
  A.~Angelova, X.~Zhai, N.~Houlsby, and R.~Soricut.
\newblock Pali: A jointly-scaled multilingual language-image model.
\newblock In {\em Proceedings of the eleventh International Conference on
  Learning Representations}, 2023.

\end{thebibliography}

\newpage

\appendix

\section{Test Accuracy plots}\label{sec_res_of_results}

\begin{figure}[hp]
    \includegraphics[width=0.9\linewidth]{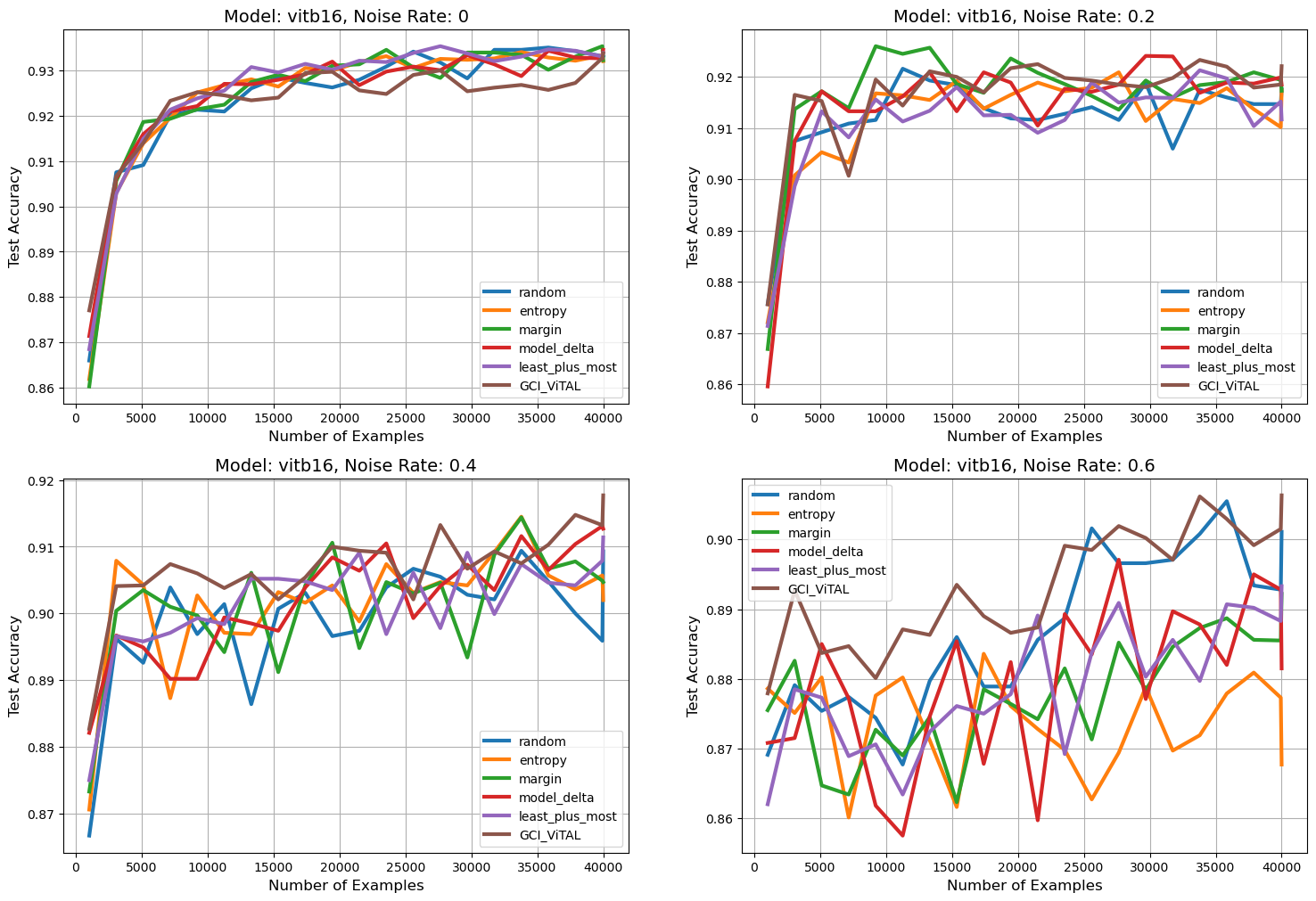} 
  \caption{Test performances of ViT transformer on CIFAR10. The graph compares each AL strategy under different label noise rates, after each round of labeling and retraining.  Under high noise rates, GCI-ViTAL performs better in both low and high labeling budget cases.}
  \label{fig:vit_cifar10_results}
\end{figure}

\begin{figure}[!hp]
    \includegraphics[width=0.9\linewidth]{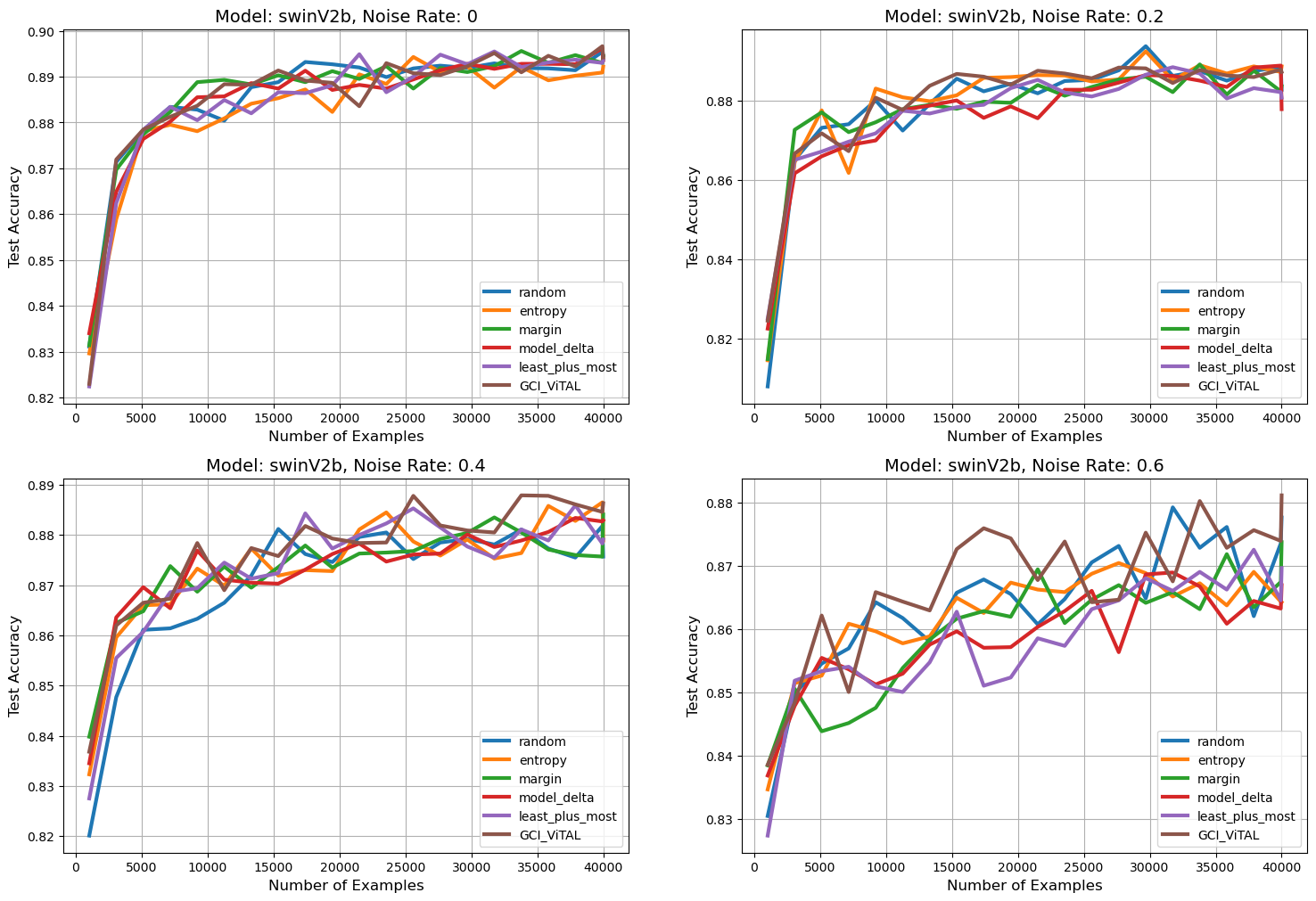} 
  \caption{Test performances of Swin transformer on CIFAR10. GCI-ViTAL performs better than other strategies under $60\%$ label noise in both low and high labeling budget cases.}
  \label{fig:swin_cifar10_results}
\end{figure}

\begin{figure}[!hp]
    \includegraphics[width=0.9\linewidth]{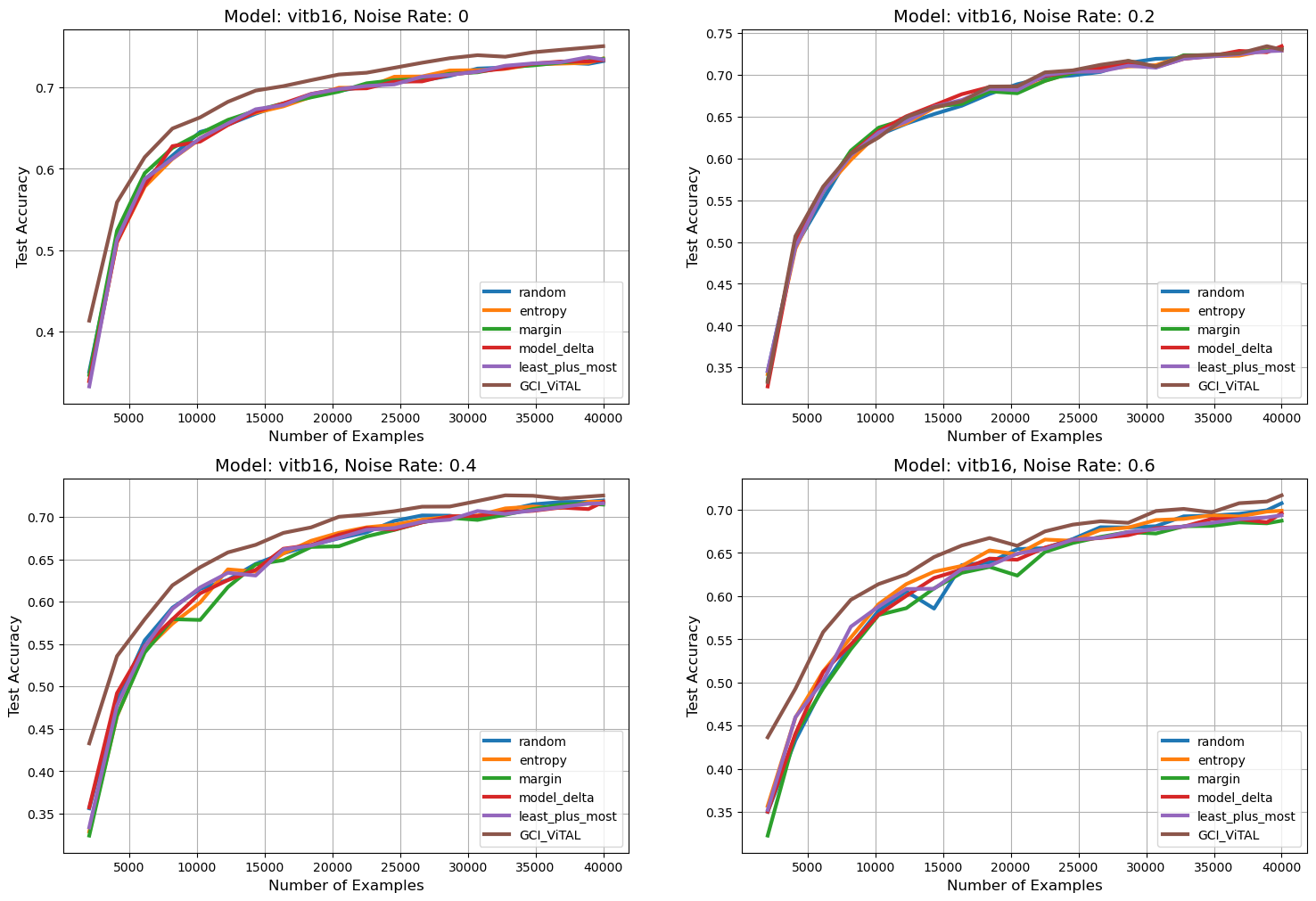}
    \caption{Test performances of ViT on CIFAR100. GCI-ViTAL performs better than other strategies with the increasing number of classes and high label noise rates due to being grounded on the C-Core attention vectors.}
    \label{fig:vit_cifar100_results}
\end{figure}

\begin{figure}[hp]
    \includegraphics[width=0.9\linewidth]{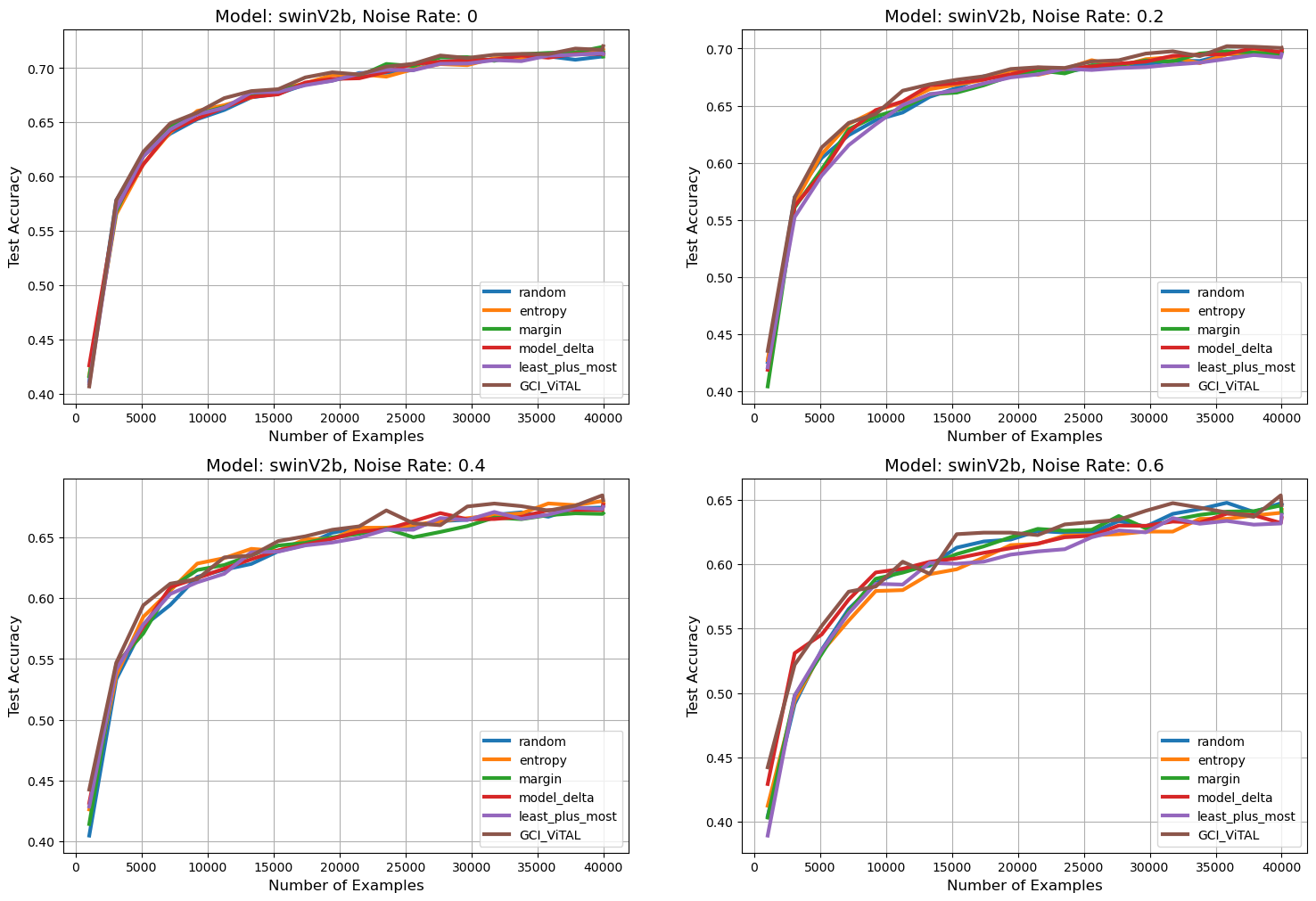} 
  \caption{Test performances of Swin transformer on CIFAR100.}
   \label{fig:swin_cifar100_results}
\end{figure}

\begin{figure}[hp]
    \includegraphics[width=0.9\linewidth]{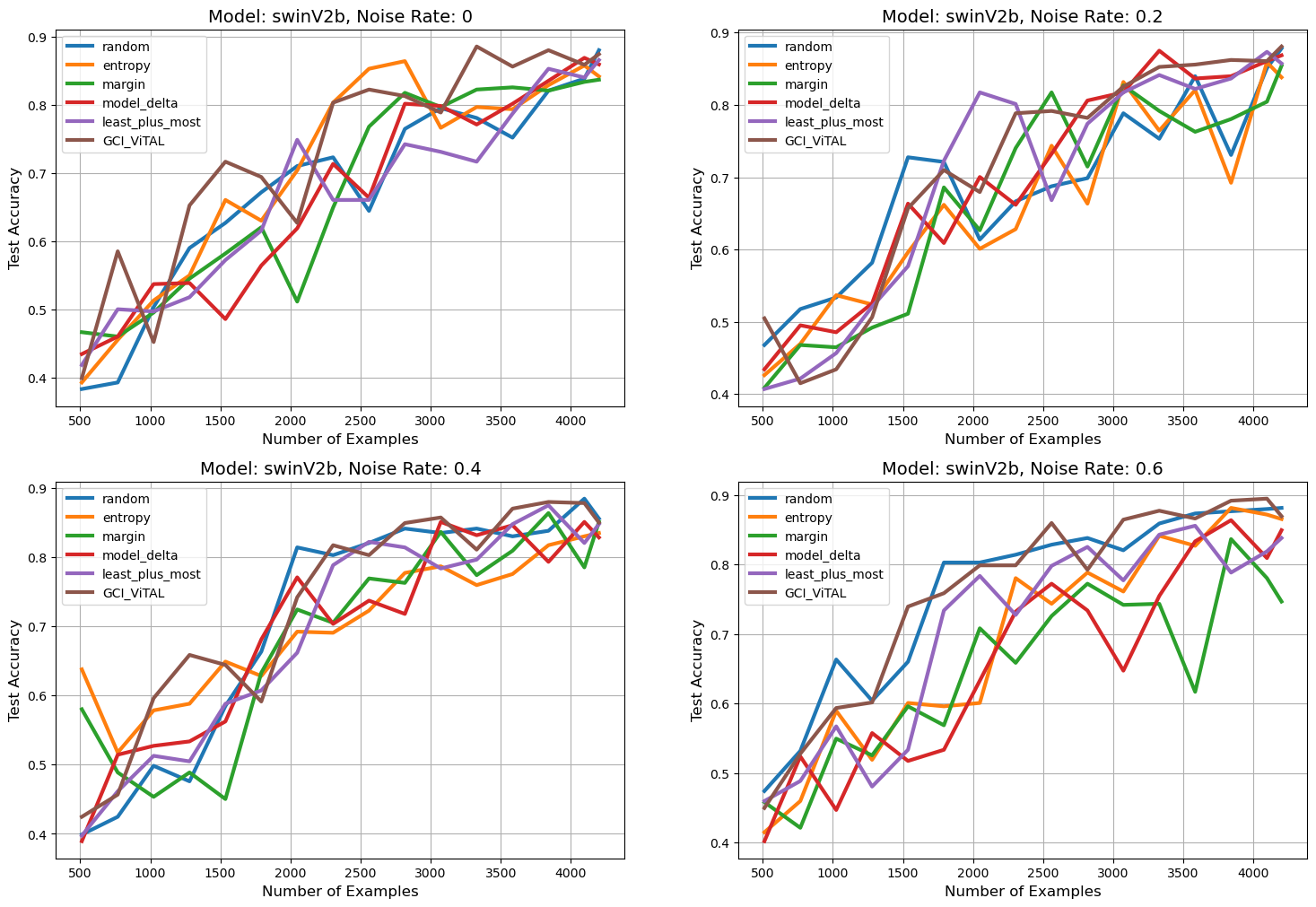}
  \caption{Test performances of Swin transformer on the Chest X-ray dataset.}
  \label{fig:vit_chest_xray_results}
\end{figure}

\begin{figure}[hp]
    \includegraphics[width=0.9\linewidth]{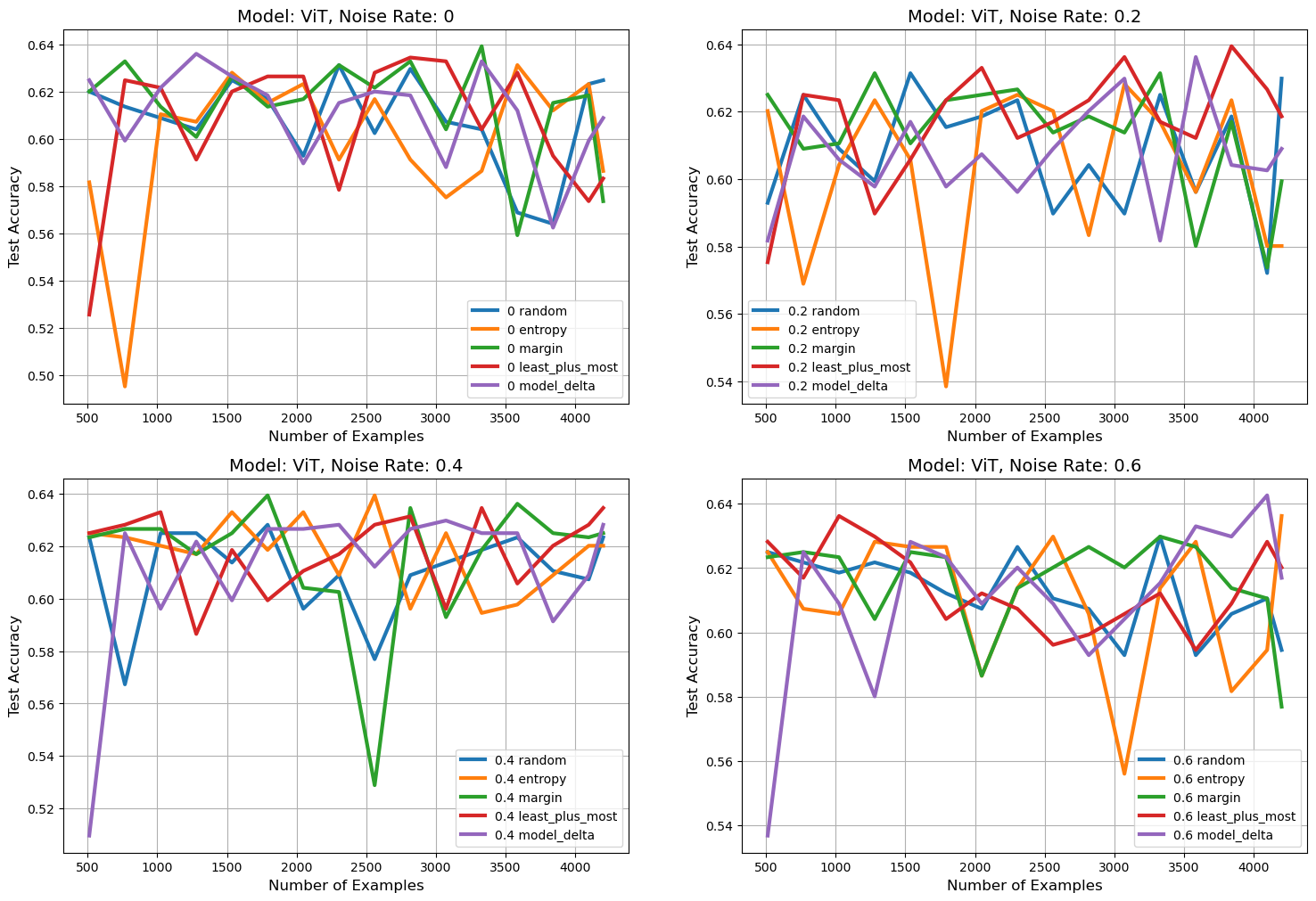} 
  \caption{Test performances of ViT transformer on the Chest X-ray dataset.}
\label{fig:swin_chest_xray_results}
\end{figure}

\begin{figure}[hp]
    \includegraphics[width=0.9\linewidth]{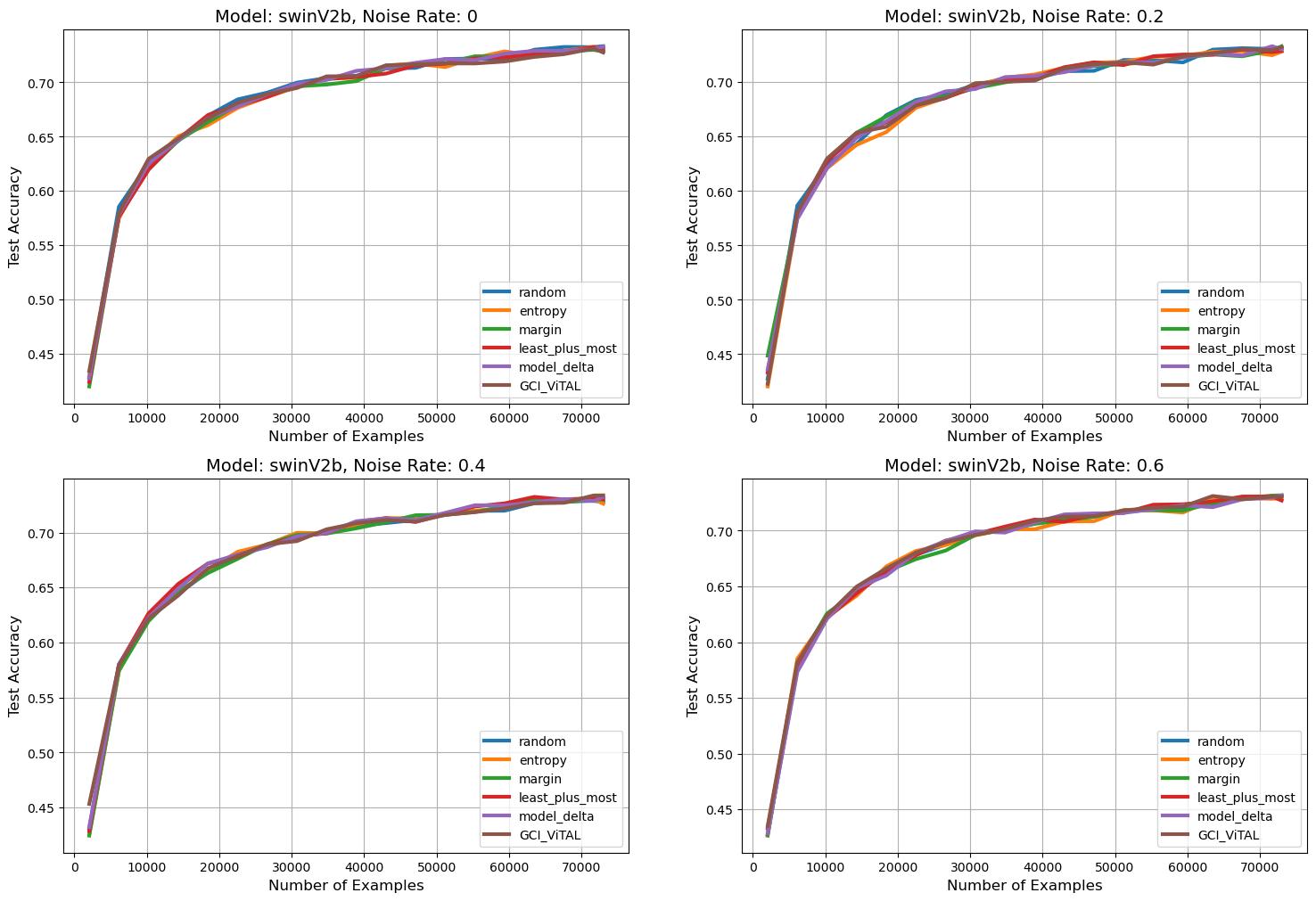}
  \caption{Test performances of ViT transformer on the more complex Food101 dataset.}
  \label{fig:swin_food101_results}
\end{figure}

\begin{figure}[htp]
    \includegraphics[width=0.9\linewidth]{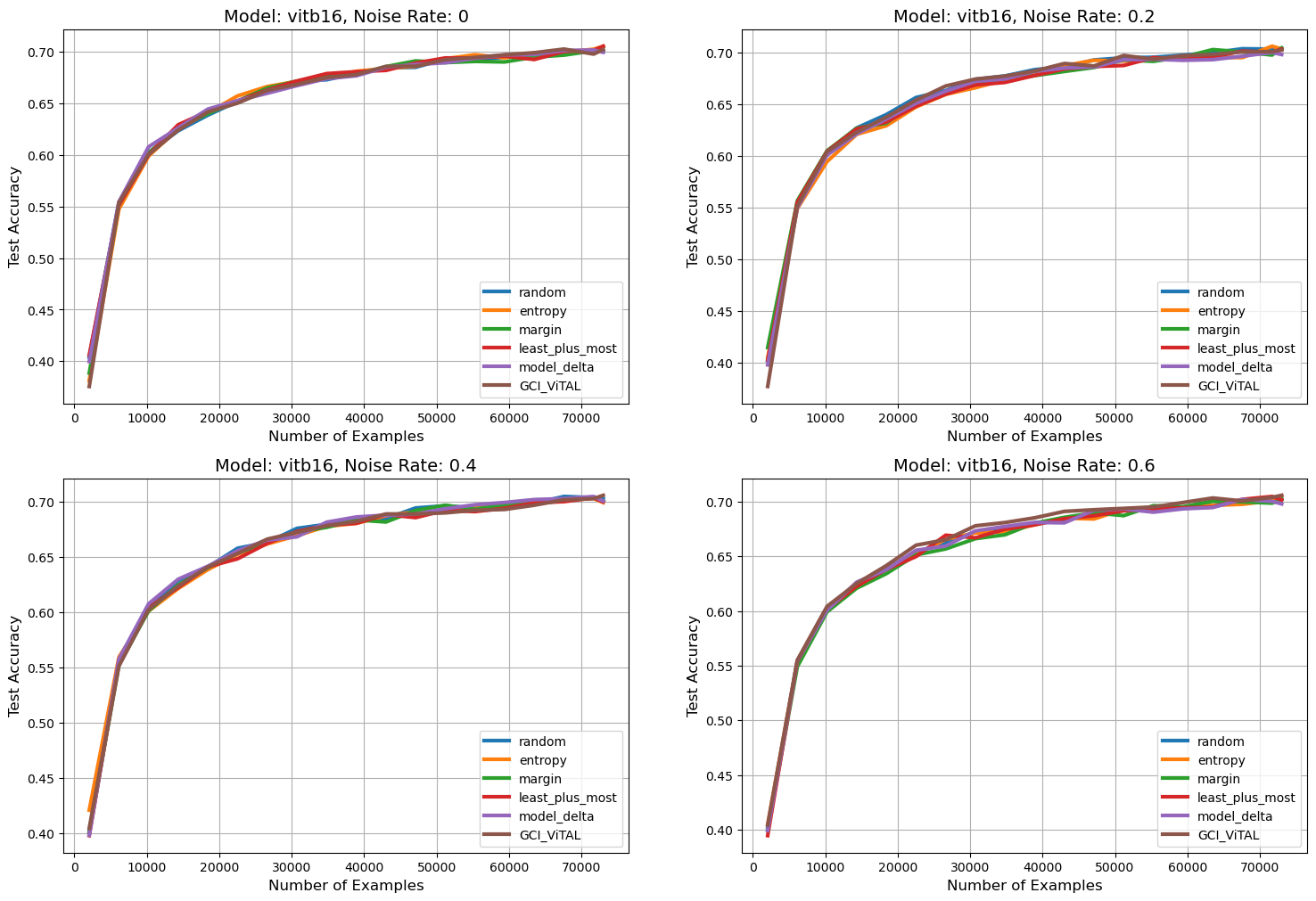} 
  \caption{Test performances of ViT transformer on the more complex Food101 dataset.}
\label{fig:vit_food101_results}
\end{figure}

\end{document}